\documentclass{article}

\PassOptionsToPackage{numbers,square}{natbib}

\usepackage[final]{neurips_2019}

\usepackage[utf8]{inputenc} 
\usepackage[T1]{fontenc}    
\usepackage{hyperref}       
\usepackage{url}            
\usepackage{booktabs}       
\usepackage{amsfonts}       
\usepackage{nicefrac}       
\usepackage{microtype}      
\usepackage{multirow}
\usepackage{subcaption}
\usepackage{xcolor}
\usepackage{amsmath}
\usepackage{amssymb}
\usepackage{graphicx}
\usepackage{floatrow}
\usepackage{enumitem}
\DeclareFloatFont{tiny}{\tiny}
\renewcommand{\footnotesize}{\scriptsize}

\newif\ifshowcomments
\showcommentsfalse

\ifshowcomments
\newcommand{\pradeep}[1]{{\color{cyan}{[Pradeep: #1]}}}
\newcommand{\aditya}[1]{{\color{violet}{[Aditya: #1]}}}
\else
\newcommand{\pradeep}[1]{}
\newcommand{\aditya}[1]{}
\fi

\title{How Powerful Are Randomly Initialized Pointcloud Set Functions?}

\author{
  Aditya Sanghi and Pradeep Kumar Jayaraman\\
  Autodesk Research, Toronto\\
  \texttt{\{aditya.sanghi, pradeep.kumar.jayaraman\}@autodesk.com}
}

\begin{document}

\maketitle

\begin{abstract}
  We study random embeddings produced by untrained neural set functions, and show that they are powerful representations which well capture the input features for downstream tasks such as classification, and are often linearly separable.
We obtain surprising results that show that random set functions can often obtain close to or even better accuracy than fully trained models.
We investigate factors that affect the representative power of such embeddings quantitatively and qualitatively.
\end{abstract}

\section{Introduction}
\label{sec:introduction}
Pointclouds are ubiquitious set-based data representations in computer vision, and numerous permutation-invariant neural networks exist in literature for processing them~\cite{qi2016:pointnet,zaheer2017:deepsets}.
In this work, we study such neural networks from a very different perspective---we randomly initialize the neural network and do not perform any training, thereby treating these networks as random nonlinear projection functions which transform the input features into latent random embeddings.
These random embeddings can be used for downstream applications, e.g. classification, and even serve as strong baselines.

\paragraph{Related Work.} This is motivated by several theoretical and practical works in the literature: Cover's theorem~\cite{cover65:theorem} states that non-linearly projecting data into high dimensional spaces improves their chances of being linearly separable.
\citet{he2016:randomvis} use randomly initialized convolutional neural networks to separately explore the importance of network structure, as opposed to training.
\citet{ulyanov2017:deepimageprior} successfully use randomly initialized neural networks as a prior for inverse image problems.
\citet{gaier2019:weightagnostic} search for neural networks architectures that are agnostic to the choice of weights.
\citet{wieting2019:randomword} employ random neural networks to compute embeddings for sentence classification.
\citet{frankle2018lottery} introduce the lottery ticket hypothesis which conjectures that randomly initialized feedforward networks contain certain sub-networks, which when trained in isolation perform comparably to the original network. 

\paragraph{Contributions.} In this work, we study randomly initialized set functions, particularly for pointclouds in a similar vein, and make the following contributions:
First, we experiment with well known set-based neural networks, but without any training.
Surprisingly, we attain close to state-of-the-art classification accuracy on MNIST, and ModelNet40, and find that the architecture plays an important role even with untrained networks.
In detail, we find that the architecture of random networks directly affects downstream classification accuracy, normalization helps embedding the input features in a better linearly separable space, and alignment of pointclouds in existing datasets is an important factor affecting the performance of both random and trained networks.
Second, we perform qualitative experiments using unsupervised learning techniques to study what the random embeddings represent.
Strikingly, we find that the random embeddings capture important global features from the input pointclouds, and preserve the per-class clustering to a large extent.

\section{Overview}
\label{sec:overview}

Our main observation in this work is that simple models such as linear classifiers perform surprisingly well on standard pointcloud datasets if trained on embeddings produced by untrained, randomly initialized set-based neural networks.
Given a pointcloud $\{\mathbf{x}_i \ | \ \mathbf{x}_i \in \mathbb{R}^3, \ i \in 1, \ldots, n \}$, our goal is to generate an $m$-dimensional embedding $\mathbf{e}_i \in \mathbb{R}^m$.
To generate $\mathbf{e}_i = f(\mathbf{x_i})$, we consider different classes of set functions $f$ parameterized by randomly initialized neural networks of different architectures.
Once we have the embeddings, we can treat them as features to train simple classifiers.
Surprisingly, we observe that these random embeddings are powerful features and often even linearly separable yielding very good accuracy on downstream tasks.
Based on this, we investigate several factors quantitatively~(Section \ref{sec:factors}) and qualitatively (Section~\ref{sec:visualization}) to understand how powerful random embeddings are and what information they capture.

\section{Factors Affecting Separability of Embeddings}
\label{sec:factors}

\paragraph{Architecture}
\label{ssec:factors:architecture}
We explore four architectures (we only consider the set function) for producing point cloud embeddings: (1) a linear model with 1D per-point convolutions followed by maxpooling for permutation invariance and no normalization (NN) called LinSet-NN, (2) a variation of LinSet with InstanceNorm~\cite{Ulyanov2016:instancenorm} called LinSet, (3) PointNet~\cite{qi2016:pointnet} (without spatial transformer networks), and (4) DeepSets~\cite{zaheer2017:deepsets}.
All architectures are made to produce pointcloud embeddings of $1024$ dimensions.
Better embedding dimensions can also be chosen according to \citet{wagstaff2019limitations}.
We initialize the parameters of these models with \citet{Glorot2010:weightinit}'s method.
Other initialization methods like \citet{he2015:weightinit}'s, or sampling from uniform or normal distributions slightly decreased the accuracy in experiments.
We trained embeddings produced from these random neural networks on:
\begin{enumerate}[leftmargin=*]
\item \textsc{LinClf}, a 1-layer linear classifier: $\text{FC}(1024, \#\text{classes})$, where $\text{FC}$ is a fully connected layer.
\item \textsc{NonLinClf}, a 3-layer non-linear classifier: $\text{FC}(1024, 512) \rightarrow \text{BN} \rightarrow \text{LeakyReLU} \rightarrow \text{Dropout}(0.8) \rightarrow \text{FC}(512, 256)\rightarrow \text{BN} \rightarrow \text{LeakyReLU} \rightarrow \text{Dropout}(0.8) \rightarrow \text{FC}(256,\#\text{classes})$
\end{enumerate}

We consider two datasets for the problem of point cloud classification: ModelNet40, and a pointcloud version of MNIST denoted MNIST-PC, generated by rejection sampling of a fixed number of points (512 in our experiments) from the white pixels.
We train the classifiers with backpropagation using the cross entropy loss, and the ADAM optimizer with the following hyperparameters: learning rate $0.001$, momentum $0.9$, mini-batch size $32$ for $5000$ epochs.
Note that it only takes $\sim 2$--$3$min. for computing the embeddings, training and validation on a Tesla K40c.
The results are shown in Table~\ref{tab:architecture}.

\begin{table}
  \caption{Pointcloud classification with random neural networks: Mean and standard deviation of test accuracy (\%) for different architectures computed over $5$ runs.}
  \label{tab:architecture}
  \centering
  \begin{tabular}{lllcc}
    \toprule
    \multirow{2}{*}{Training}         & \multirow{2}{*}{Architecture}  & \multirow{2}{*}{Classifier}   & \multicolumn{2}{c}{Test Accuracy (\%)}\\
                                      &                                &                               & MNIST-PC                 & ModelNet40 \\
    \midrule
    \multirow{8}{*}{Classifier only}  & \multirow{2}{*}{LinSet-NN}     & \textsc{LinClf}               & 79.407 $\pm$ 0.049       & 68.417 $\pm$ 0.271 \\
                                      &                                & \textsc{NonLinClf}            & 87.989 $\pm$ 0.024       & 81.177 $\pm$ 0.180 \\
                                      & \multirow{2}{*}{LinSet}        & \textsc{LinClf}               & 84.966 $\pm$ 0.363       & 75.373 $\pm$ 0.255 \\
                                      &                                & \textsc{NonLinClf}            & 92.734 $\pm$ 0.055       & 83.482 $\pm$ 0.124 \\
                                      & \multirow{2}{*}{PointNet}      & \textsc{LinClf}               & 95.204 $\pm$ 0.228       & 83.929 $\pm$ 0.334 \\
                                      &                                & \textsc{NonLinClf}            & {\bf 97.752 $\pm$ 0.072} & {\bf 85.235 $\pm$ 0.666} \\
                                      & \multirow{2}{*}{DeepSets\footnotemark}
                                                                       & \textsc{LinClf}               & 71.817 $\pm$ 0.156       & 61.940 $\pm$ 0.277 \\
                                      &                                & \textsc{NonLinClf}            & 91.546 $\pm$ 0.359       & 82.597 $\pm$ 0.245 \\
    \midrule
    \multirow{2}{*}{Full}             & \multicolumn{2}{c}{FoldingNet~\cite{yang2018:foldingnet}}      & N/A                      & 88.44 \\
                                      & \multicolumn{2}{c}{PointNet~\cite{qi2016:pointnet}}            & N/A                      & {\bf 89.20} \\
                                      & \multicolumn{2}{c}{3D-GAN~\cite{jiajun2016:3dgan}}             & N/A                      & 83.33 \\
                                      & \multicolumn{2}{c}{Latent-GAN~\cite{achlioptas2018:latentgan}} & N/A                      & 85.50 \\
    \bottomrule
  \end{tabular}
\end{table}
\footnotetext{The original DeepSets implementation does not include normalization layers, and this appears to affect accuracy (see Section~\ref{ssec:factors:normalization}).}
We observe that all models generally perform surprisingly well, given that the neural networks are completely untrained.
In particular, we find PointNet to give close to state-of-the-art results, and sometimes even higher than trained autoencoders that use their learnt representation for classification.
We next investigate other factors that influence the representative power of random embeddings.


\paragraph{Normalization}
\label{ssec:factors:normalization}
We will now show that the choice of using normalization is a critical design criteria in creating these embeddings. 
To test the effect of normalization, we experiment on PointNet which already has BatchNorm (BN)~\cite{ioffe2015batch}. 
We additionally replace BatchNorm with Instance Normalization (IN)~\cite{Ulyanov2016:instancenorm}, Layer Normalization (LN)~\cite{ba2016layer}, and no normalization (NN).
We do not include any learnable parameters in the normalization modules.
Several interesting observations can be made from the results in Table~\ref{tab:normalization} which were computed over $5$ runs on the ModelNet40 dataset.
Firstly, despite not being trained, the embeddings produced by the model perform remarkably well on the downstream classification task.
Secondly, the separability of embeddings is affected by the choice of using normalization---with normalization, a linear classifier is enough to get high accuracy; without normalization, the accuracy drops with \textsc{LinClf}, and \textsc{NonLinClf} is required to obtain similar accuracy.
This suggests that neural set functions with normalization produce embeddings that are better linearly separable.
\begin{table}
  \caption{Effect of Normalization on PointNet classification accuracy (\%)}
  \label{tab:normalization}
  \centering
  \begin{tabular}{lcc}
  \toprule
  Normalization  & \textsc{LinClf}           & \textsc{NonLinClf} \\
  \midrule
  BN     & 83.580 $\pm$ 0.421        & 85.771 $\pm$ 0.126 \\
  IN     & {\bf 86.193 $\pm$ 0.232}  & {\bf 86.631 $\pm$ 0.250} \\
  LN     & 83.214 $\pm$ 0.241        & 86.315 $\pm$ 0.121 \\
  NN     & 19.927 $\pm$ 3.549        & 86.258 $\pm$ 0.376 \\
  \bottomrule
  \end{tabular}
\end{table}

\paragraph{Number of Layers}
We next investigate the relationship between number of layers  and the quality of the random embeddings produced. 
In this case, we keep adding intermediate ``MLP'' blocks to PointNet (see~\cite{qi2016:pointnet} for detail), obtain random embeddings, and train \textsc{LinClf} and \textsc{NonLinClf}.
We omit the case with $4$ MLP layers since it is equivalent to the original model.
We use models with/without InstanceNorm for the experiments (see Table~\ref{tab:layers}).
It can be observed from the results that the number of layers affects the accuracy significantly. 
For models with normalization, the classification accuracy quickly increases with number of layers and saturates after 2 layers. 
This could be related to the complexity of the dataset.
However, the accuracy decreases in the case of models without normalization.
This is a very interesting result, and suggests that increasing layers without normalization creates embeddings that are less linearly separable.

\begin{table}
  \caption{Effect of number of layers on PointNet classification accuracy}
  \label{tab:layers}
  \centering
  \begin{tabular}{ccccc}
    \toprule
    \multirow{2}{*}{\# MLP Layers}  & \multicolumn{2}{c}{MNIST-PC}  & \multicolumn{2}{c}{ModelNet40}\\
                                           & PointNet-IN                  & PointNet-NN              & PointNet-IN        & PointNet-NN\\
    \midrule
    1                               & 84.603 $\pm$ 0.289            & 80.232 $\pm$ 0.171  & 75.414 $\pm$ 0.366  & 72.232 $\pm$ 0.129 \\
    2                               & 95.994 $\pm$ 0.108            & 83.810 $\pm$ 0.933  & 86.006 $\pm$ 0.460  & 71.542 $\pm$ 0.964 \\
    3                               & 96.520 $\pm$ 0.141            & 78.948 $\pm$ 0.945  & 86.567 $\pm$ 0.367  & 57.841 $\pm$ 1.546 \\
    5                               & 96.536 $\pm$ 0.185            & 58.031 $\pm$ 3.196  & 86.445 $\pm$ 0.456  & 15.065 $\pm$ 1.977 \\
    \bottomrule
  \end{tabular}
\end{table}

\paragraph{Aligned Datasets}
\label{ssec:factors:aligned}
We observe that the standard datasets like ModelNet40 are well-aligned.
To understand how sensitive random embeddings are to alignment, we randomly rotate the dataset with arbitrary 3D rotations (see Table~\ref{tab:alignment}), and make a comparison with PointNet fully trained on the same rotated dataset.
Results indicate that alignment plays a huge role in classification accuracy in both trained and random networks.
Furthermore, this suggests that alignment might discourage the encoder from learning meaningful representations. 
It would be interesting to explore how random rotation-invariant set architectures perform in this scenario in future.

\begin{table}
  \caption{Effect of dataset alignment on PointNet-IN for classification on ModelNet40 (\%)}
  \label{tab:alignment}
  \centering
  \begin{tabular}{lcc}
    \toprule
    \multirow{2}{*}{Classifier}    & \multicolumn{2}{c}{Training} \\
                                   & Full                   & Classifier only\\
    \midrule
    \textsc{LinClf}                & 10.429 $\pm$ 5.457     & 32.695 $\pm$ 1.070\\
    \textsc{NonLinClf}             & 61.929 $\pm$ 0.453     & 37.532 $\pm$ 0.875\\
    \bottomrule
  \end{tabular}
\end{table}

\section{What do the Random Embeddings Represent?}
\label{sec:visualization}
We now apply techniques from  unsupervised learning to attempt to understand what information is contained in the random embeddings, and how generalizable is the information to other downstream tasks.
In detail, we perform clustering, train a decoder on the random embeddings, and perform t-SNE dimensionality reduction to qualitatively check if information from the pointclouds are preserved in the random embeddings.
We find that random embeddings preserve not only the clusters with respect to the data classes, but also capture global information that can be used to reconstruct the original shape to a large extent as explained below.

\paragraph{Clustering}
We apply K-Means clustering on the random embeddings generated from MNIST test set and ModelNet40 train set.
We set K as the number of classes in the dataset, and initialize the clusters using K-Means\texttt{++}.
Then, we use the Adjusted Mutual Information metric (AMI) to compare the labels assigned by the clustering with the ground truth.
The results summarized in Table~\ref{tab:clustering} suggest that random embeddings do contain class-specific information, and reinforce the idea of normalization being important even if the network is not trained.
\begin{table}
  \caption{Clustering accuracy using AMI for different architectures}
  \label{tab:clustering}
  \centering
  \begin{tabular}{lcc}
    \toprule
   Architecture   & \multicolumn{2}{c}{Datasets}\\
                  & MNIST (test set)        & ModelNet40 (train set) \\
    \midrule
    LinSet        & 0.575 $\pm$ 0.006       & 0.479 $\pm$ 0.009  \\
    LinSet-NN     & 0.370 $\pm$ 0.003       & 0.511 $\pm$ 0.003  \\
    PointNet-IN   & 0.533 $\pm$ 0.021       & 0.541 $\pm$ 0.004  \\
    PointNet-NN   & 0.336 $\pm$ 0.021       & 0.493 $\pm$ 0.008 \\
    DeepSets      & 0.525 $\pm$ 0.005       & 0.523 $\pm$ 0.005  \\
    \bottomrule
  \end{tabular}
\end{table}
\paragraph{Autoencoder on Random Projections}
To visually understand the amount of information contained in the embeddings, we trained a fully connected decoder as in~\cite{achlioptas2018:latentgan} on the random embeddings computed on the \textsc{Chairs} category in the ShapeNet training set, by minimizing the Chamfer loss with the ground truth.
During inference, we compute random embeddings from the corresponding test set, predict the 3D shape with the decoder, and compare it with the ground truth.
\newcommand{\TrLen}{2.5}
\newcommand{\TextWidthPer}{0.25}
\begin{figure}
\begin{subfigure}[b]{\TextWidthPer\textwidth}
\includegraphics[trim={\TrLen cm \TrLen cm \TrLen cm \TrLen cm},clip,width=\textwidth]{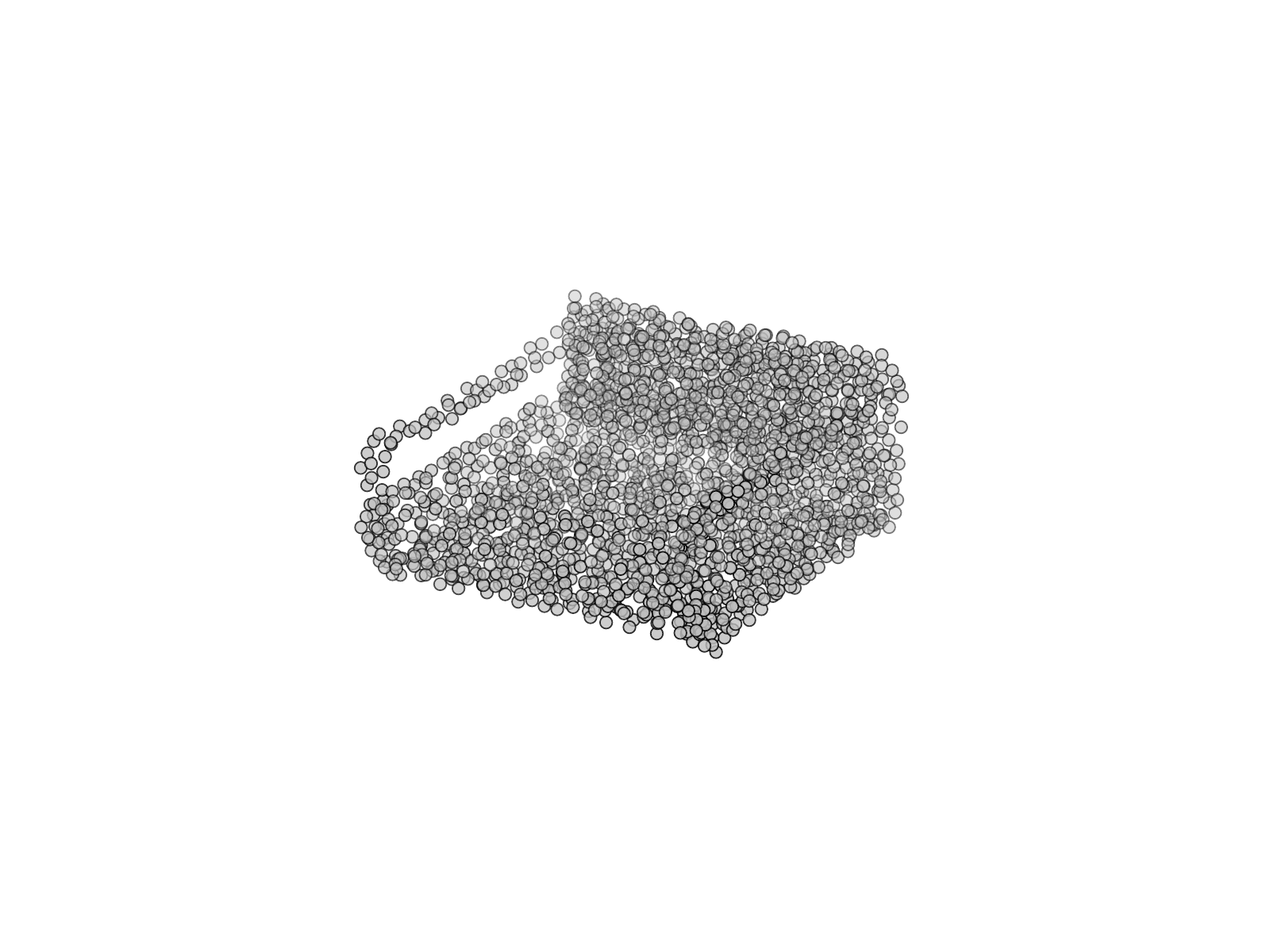}
\end{subfigure}
\begin{subfigure}[b]{\TextWidthPer\textwidth}
\includegraphics[trim={\TrLen cm \TrLen cm \TrLen cm \TrLen cm},clip,width=\textwidth]{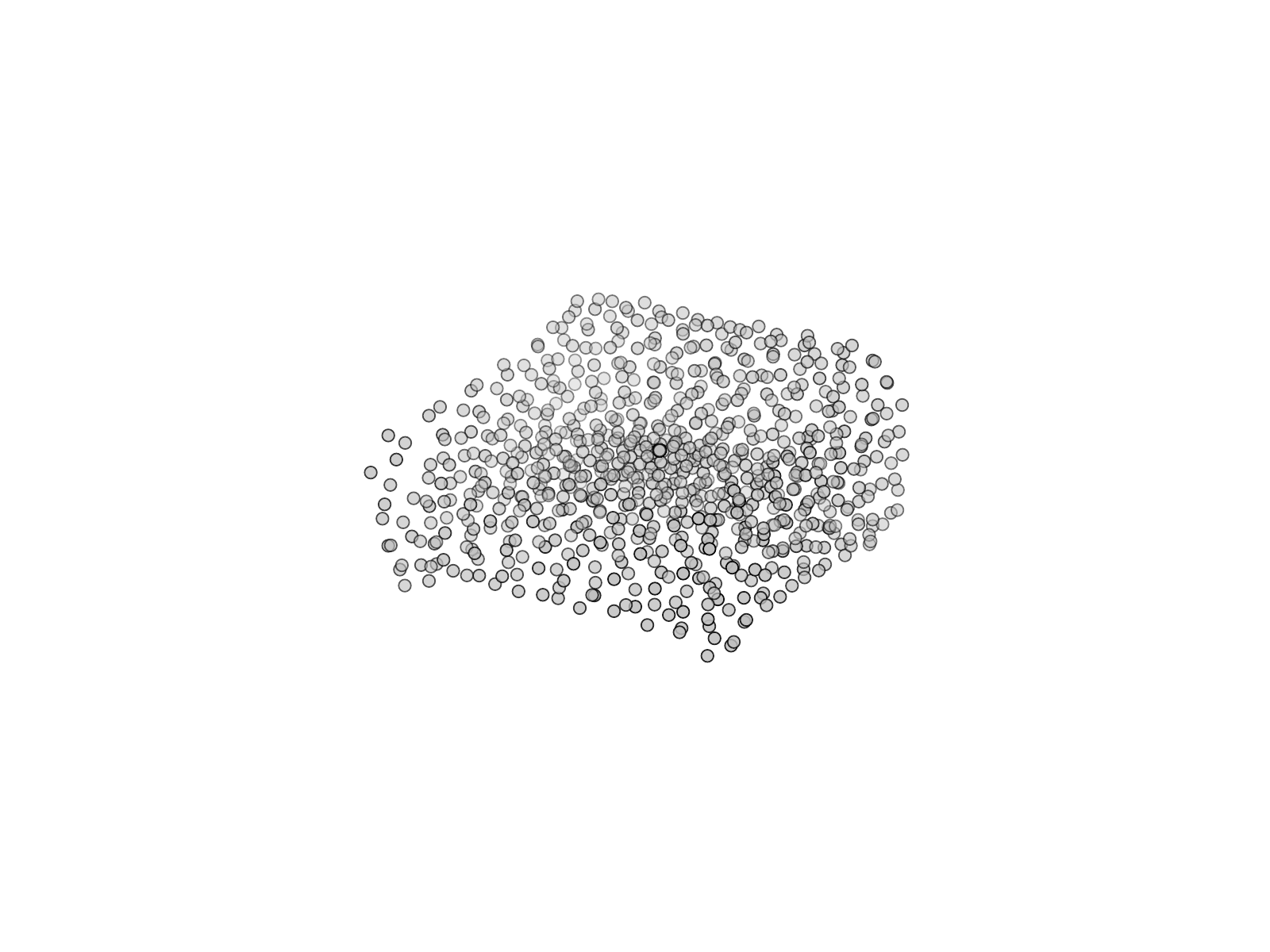}
\end{subfigure}
\begin{subfigure}[b]{\TextWidthPer\textwidth}
\includegraphics[trim={\TrLen cm \TrLen cm \TrLen cm \TrLen cm},clip,width=\textwidth]{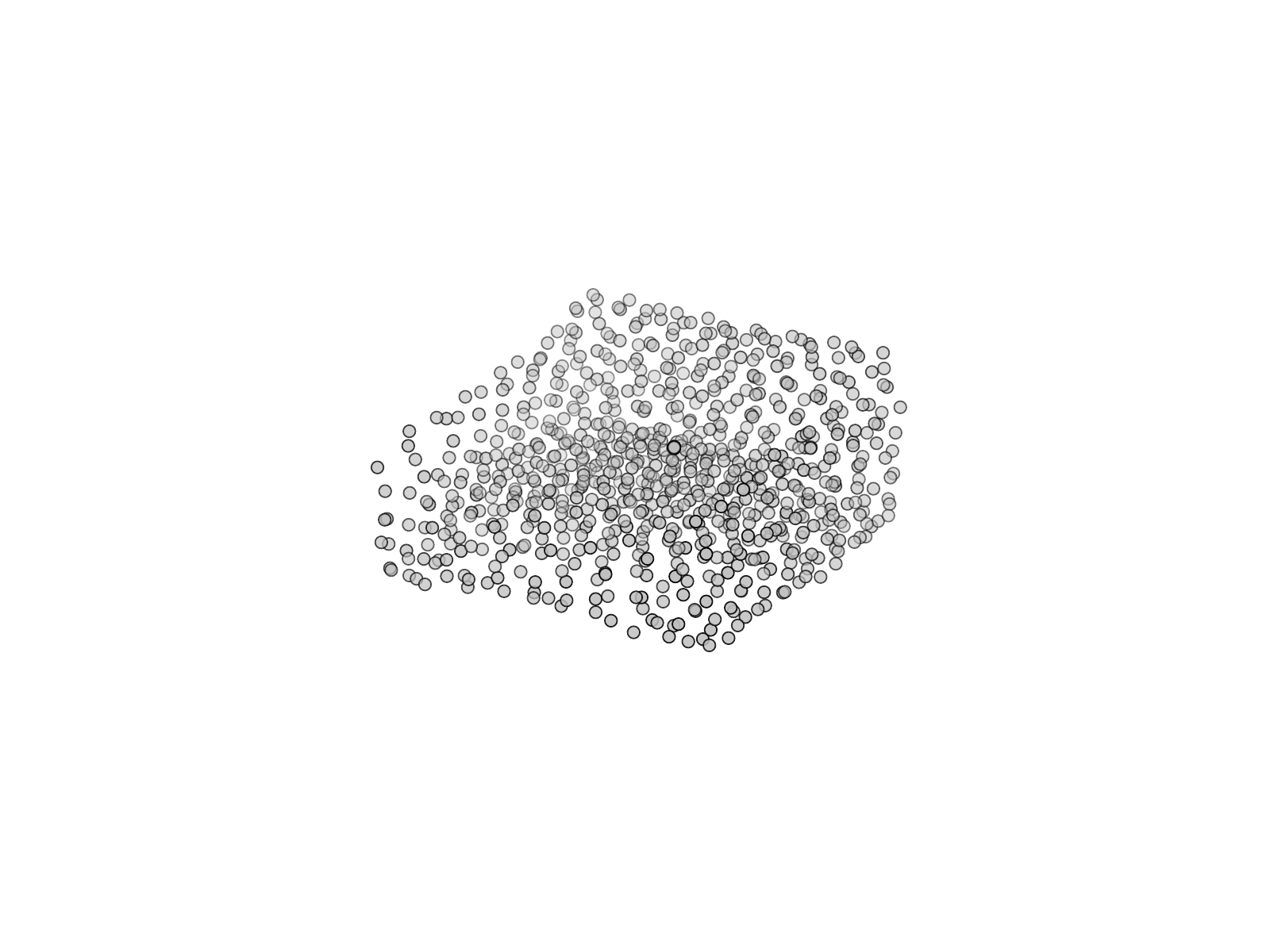}
\end{subfigure}\\
\begin{subfigure}[b]{\TextWidthPer\textwidth}
\includegraphics[trim={\TrLen cm \TrLen cm \TrLen cm \TrLen cm},clip,width=\textwidth]{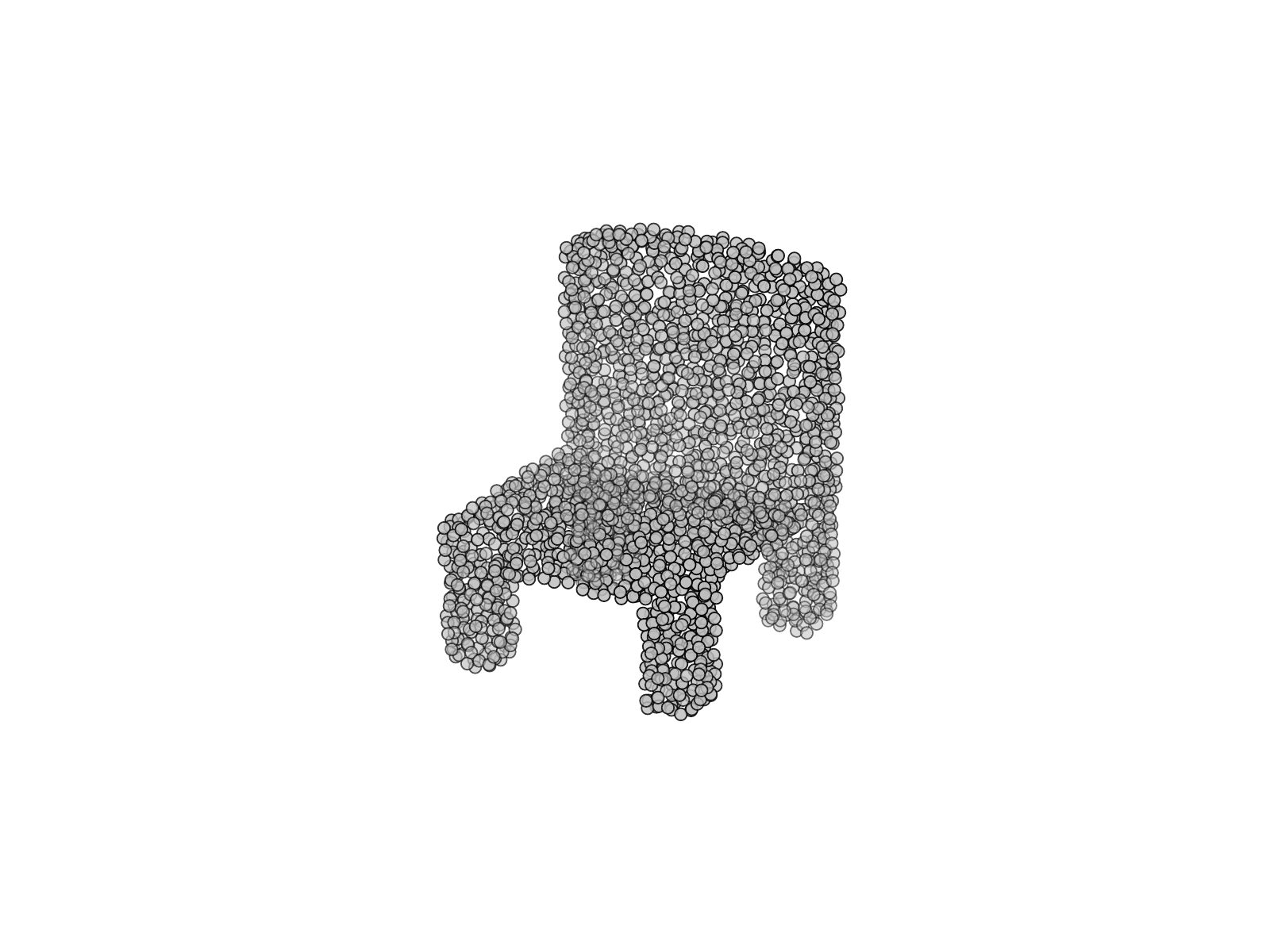}
\end{subfigure}
\begin{subfigure}[b]{\TextWidthPer\textwidth}
\includegraphics[trim={\TrLen cm \TrLen cm \TrLen cm \TrLen cm},clip,width=\textwidth]{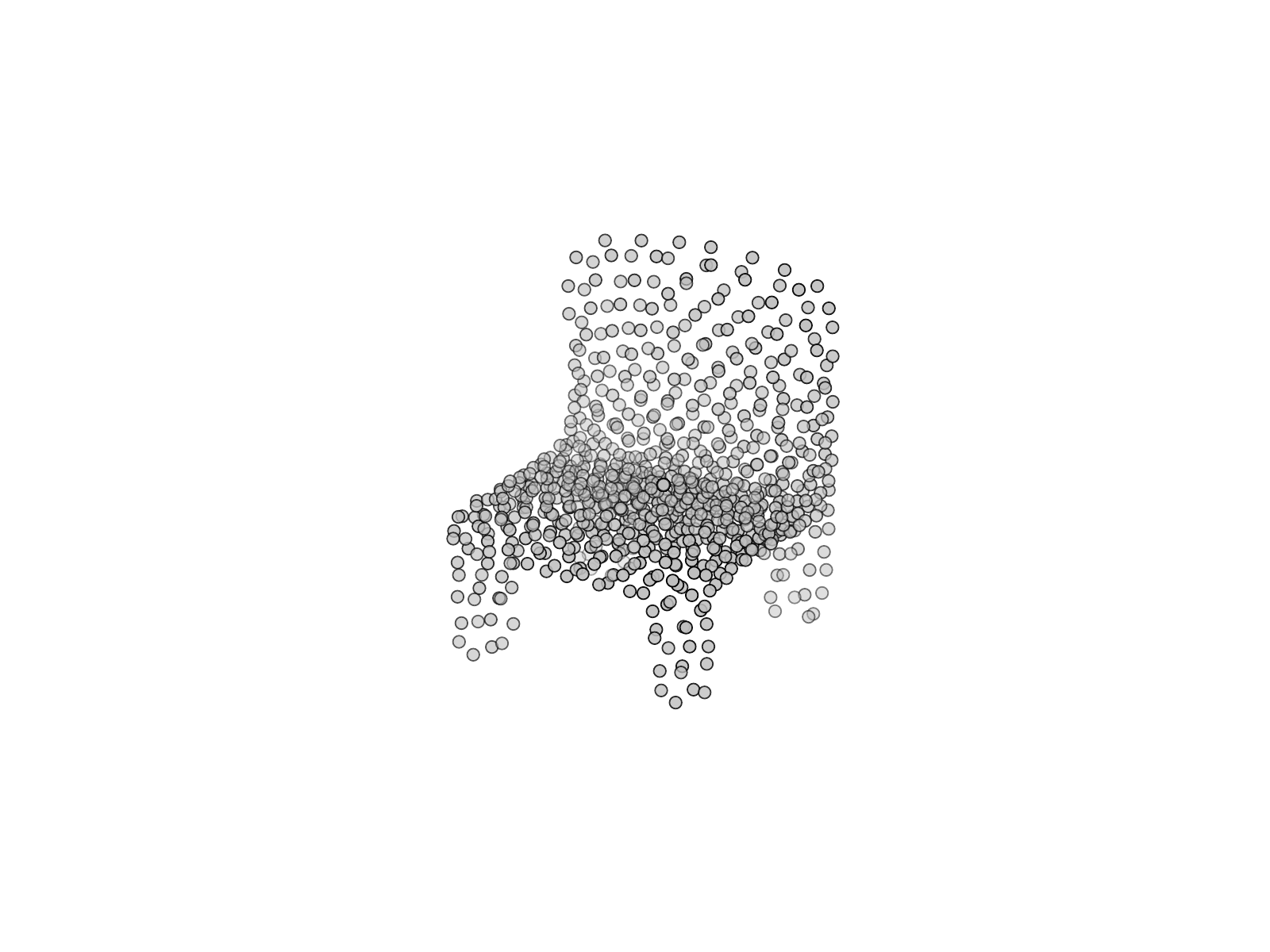}
\end{subfigure}
\begin{subfigure}[b]{\TextWidthPer\textwidth}
\includegraphics[trim={\TrLen cm \TrLen cm \TrLen cm \TrLen cm},clip,width=\textwidth]{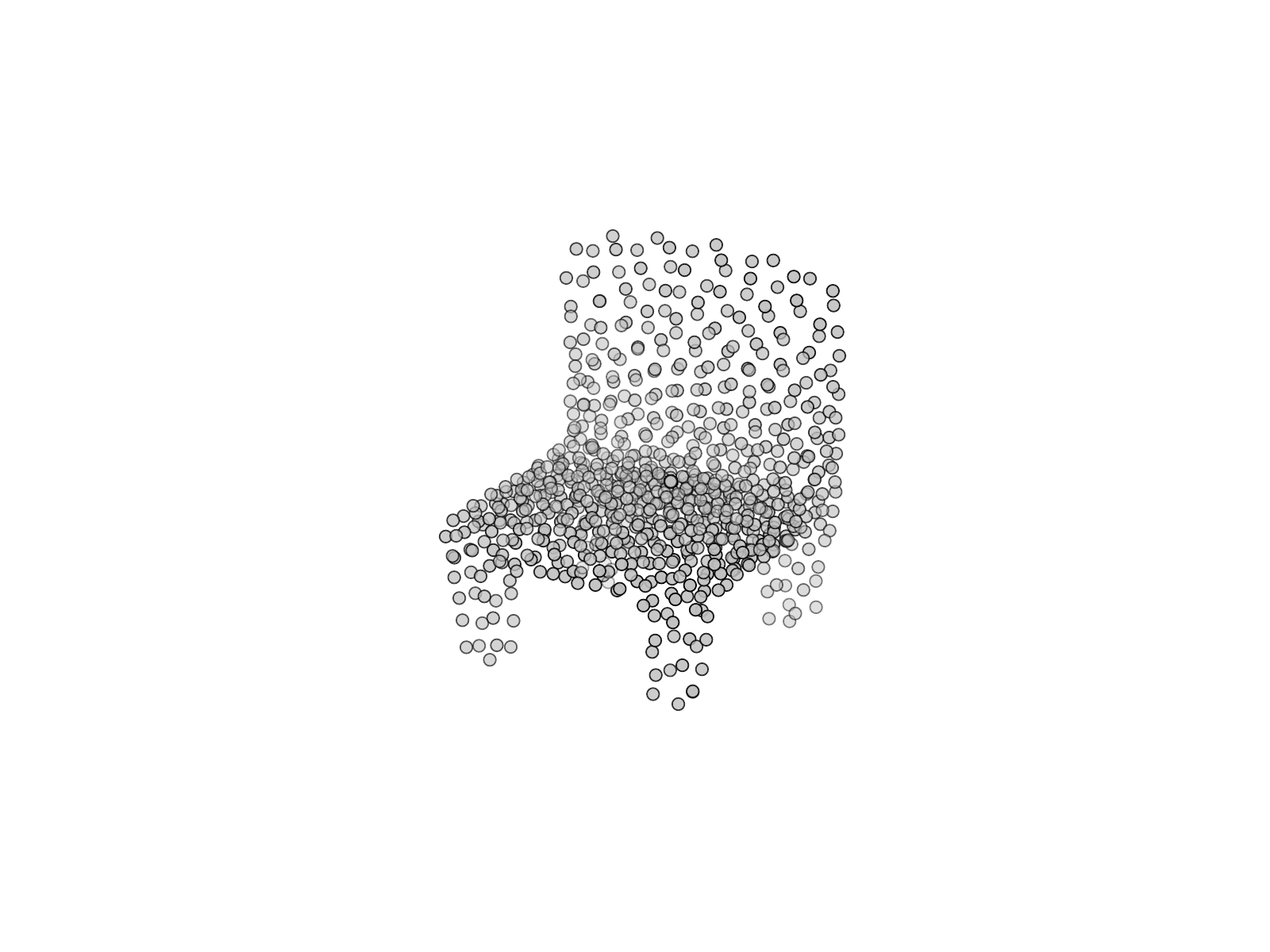}
\end{subfigure}\\
\begin{subfigure}[b]{\TextWidthPer\textwidth}
\includegraphics[trim={\TrLen cm \TrLen cm \TrLen cm \TrLen cm},clip,width=\textwidth]{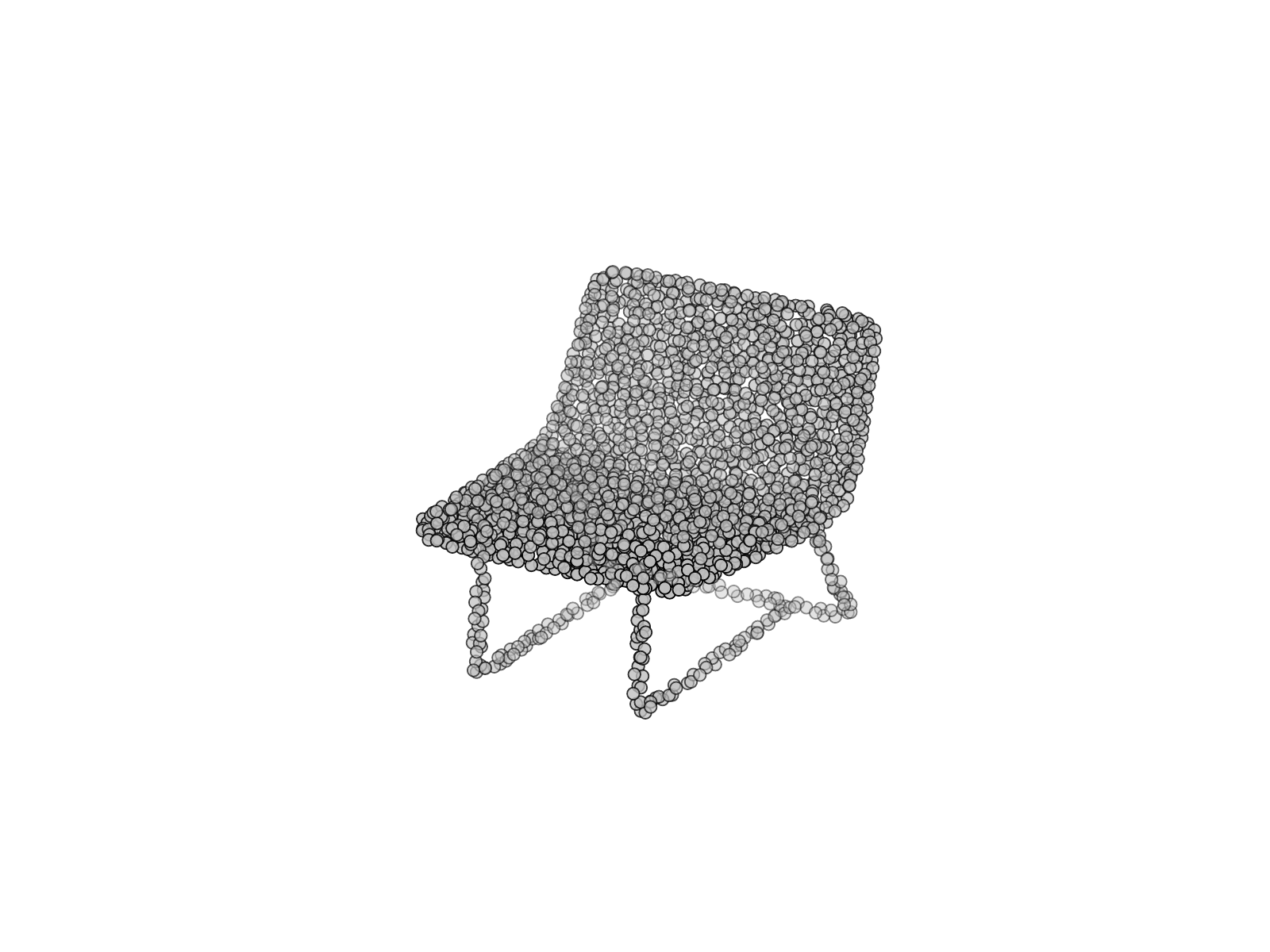}
\end{subfigure}
\begin{subfigure}[b]{\TextWidthPer\textwidth}
\includegraphics[trim={\TrLen cm \TrLen cm \TrLen cm \TrLen cm},clip,width=\textwidth]{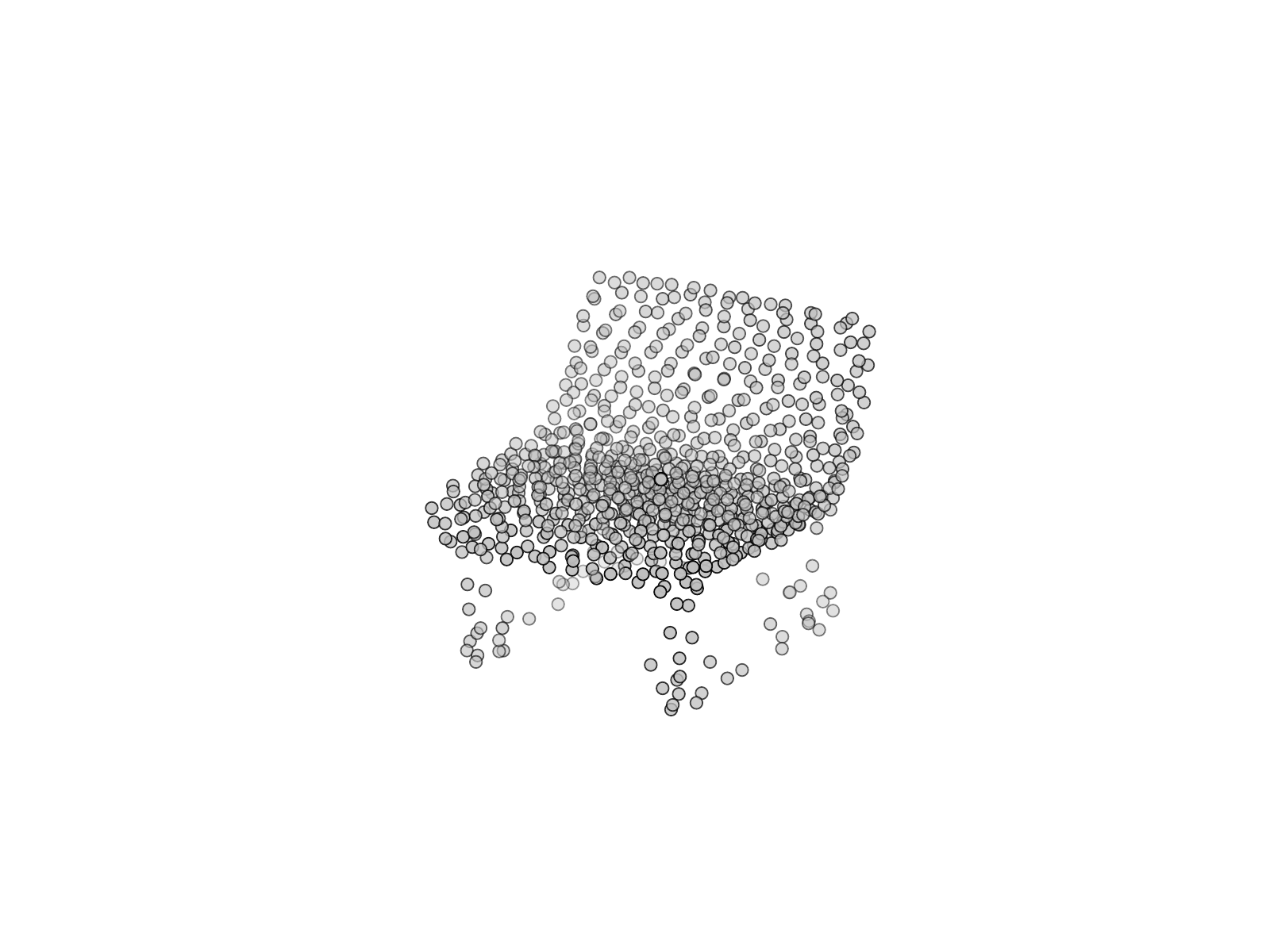}
\end{subfigure}
\begin{subfigure}[b]{\TextWidthPer\textwidth}
\includegraphics[trim={\TrLen cm \TrLen cm \TrLen cm \TrLen cm},clip,width=\textwidth]{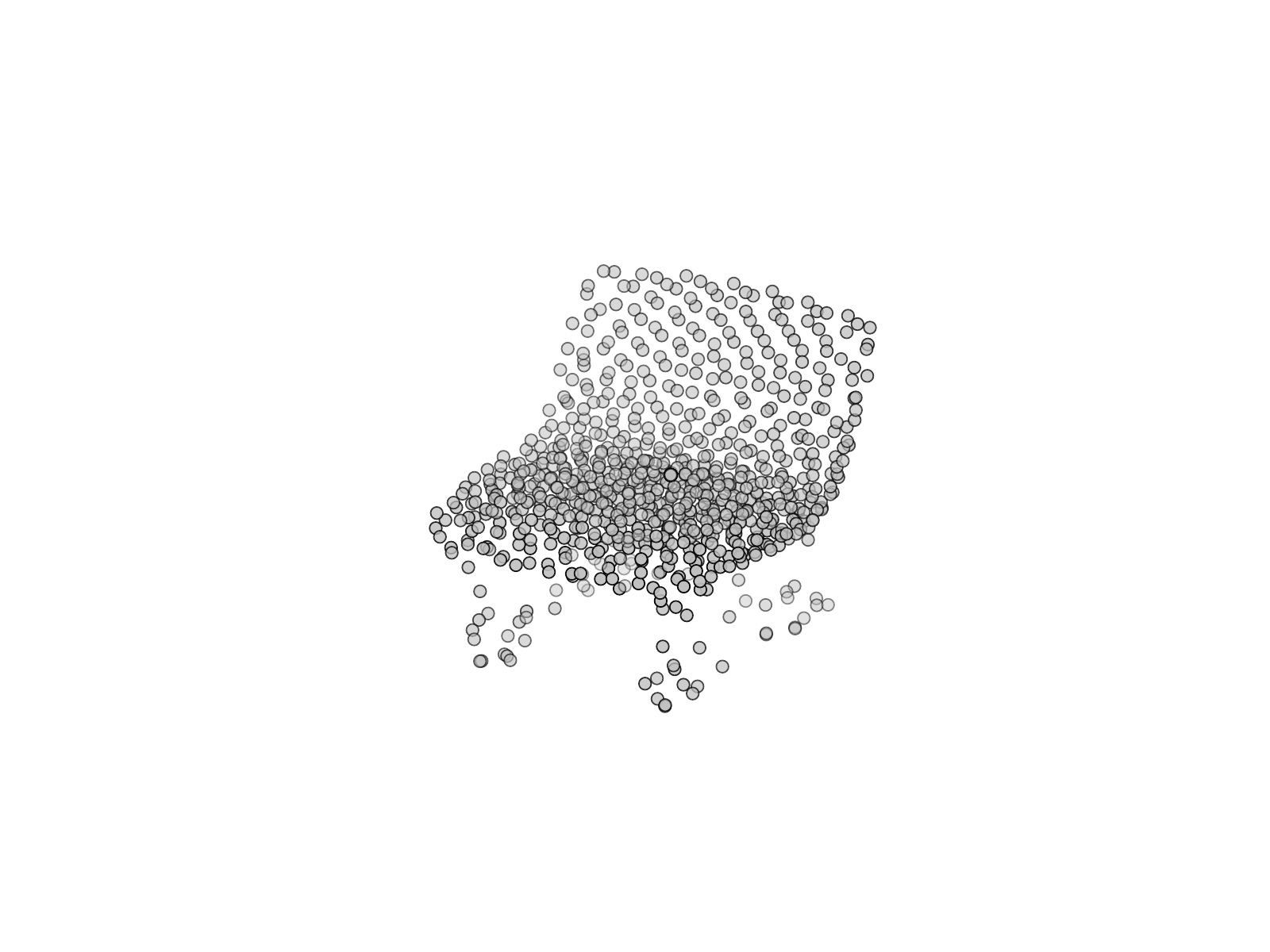}
\end{subfigure}\\
\begin{subfigure}[b]{\TextWidthPer\textwidth}
\includegraphics[trim={\TrLen cm \TrLen cm \TrLen cm \TrLen cm},clip,width=\textwidth]{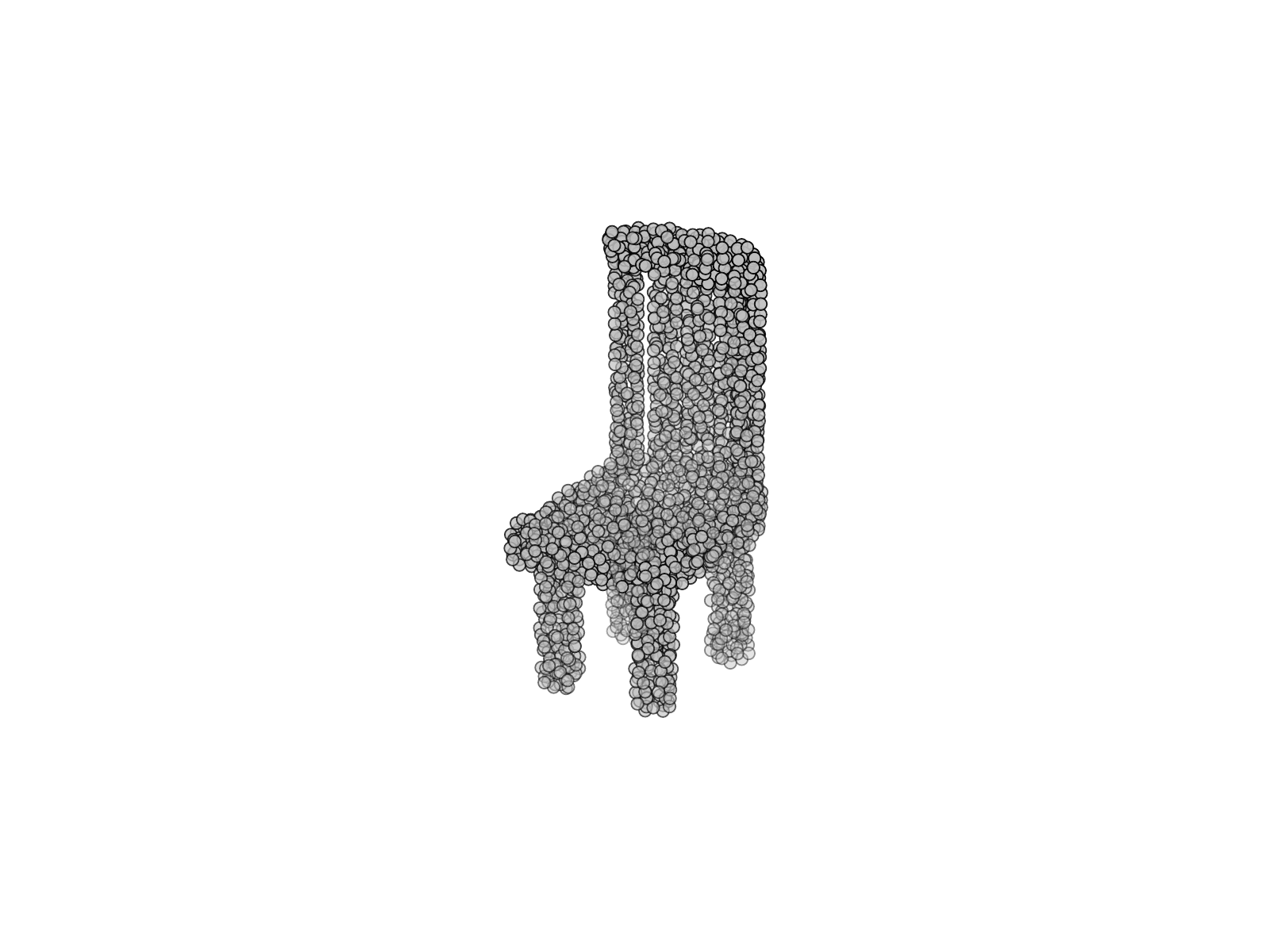}
\end{subfigure}
\begin{subfigure}[b]{\TextWidthPer\textwidth}
\includegraphics[trim={\TrLen cm \TrLen cm \TrLen cm \TrLen cm},clip,width=\textwidth]{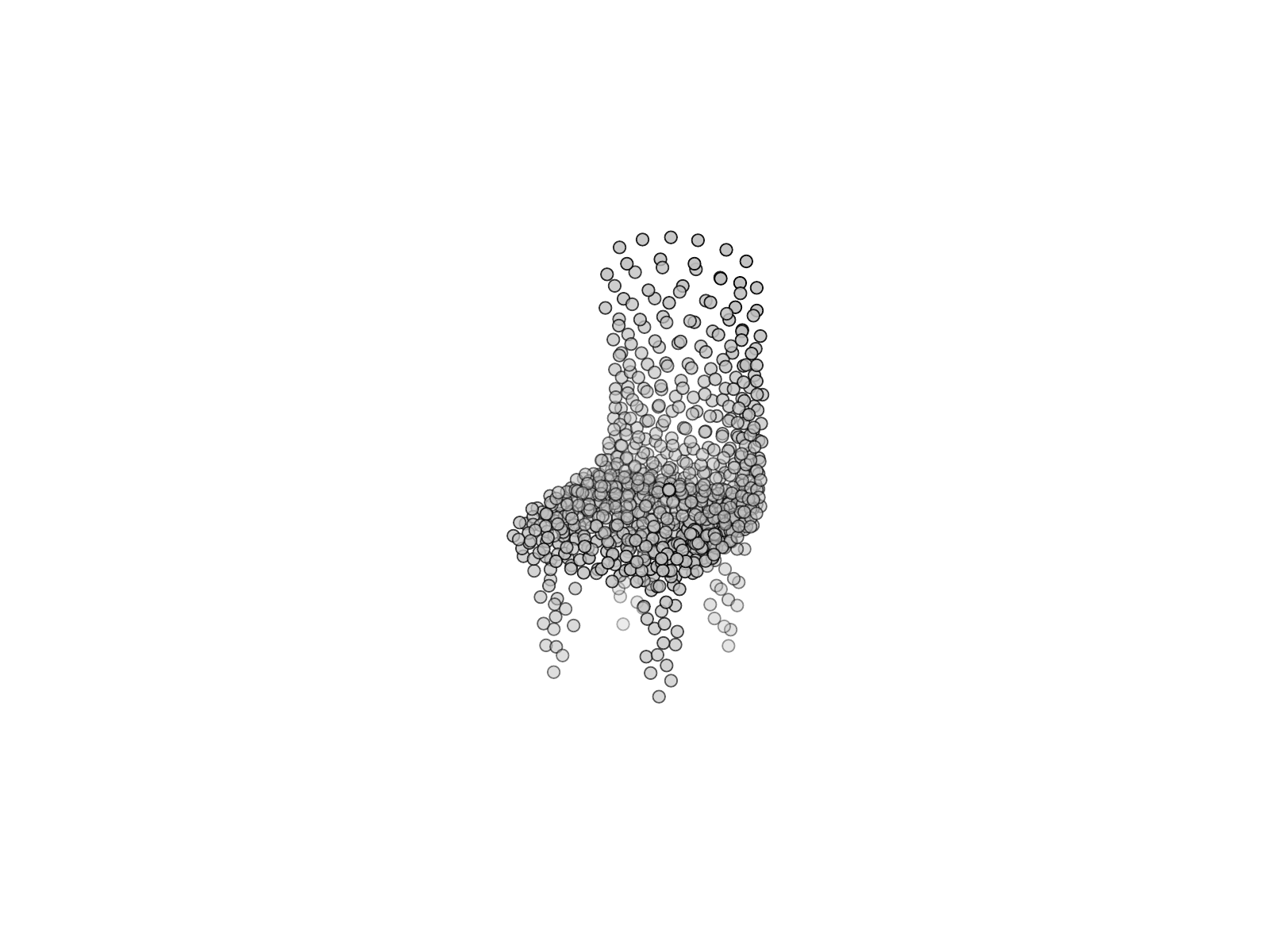}
\end{subfigure}
\begin{subfigure}[b]{\TextWidthPer\textwidth}
\includegraphics[trim={\TrLen cm \TrLen cm \TrLen cm \TrLen cm},clip,width=\textwidth]{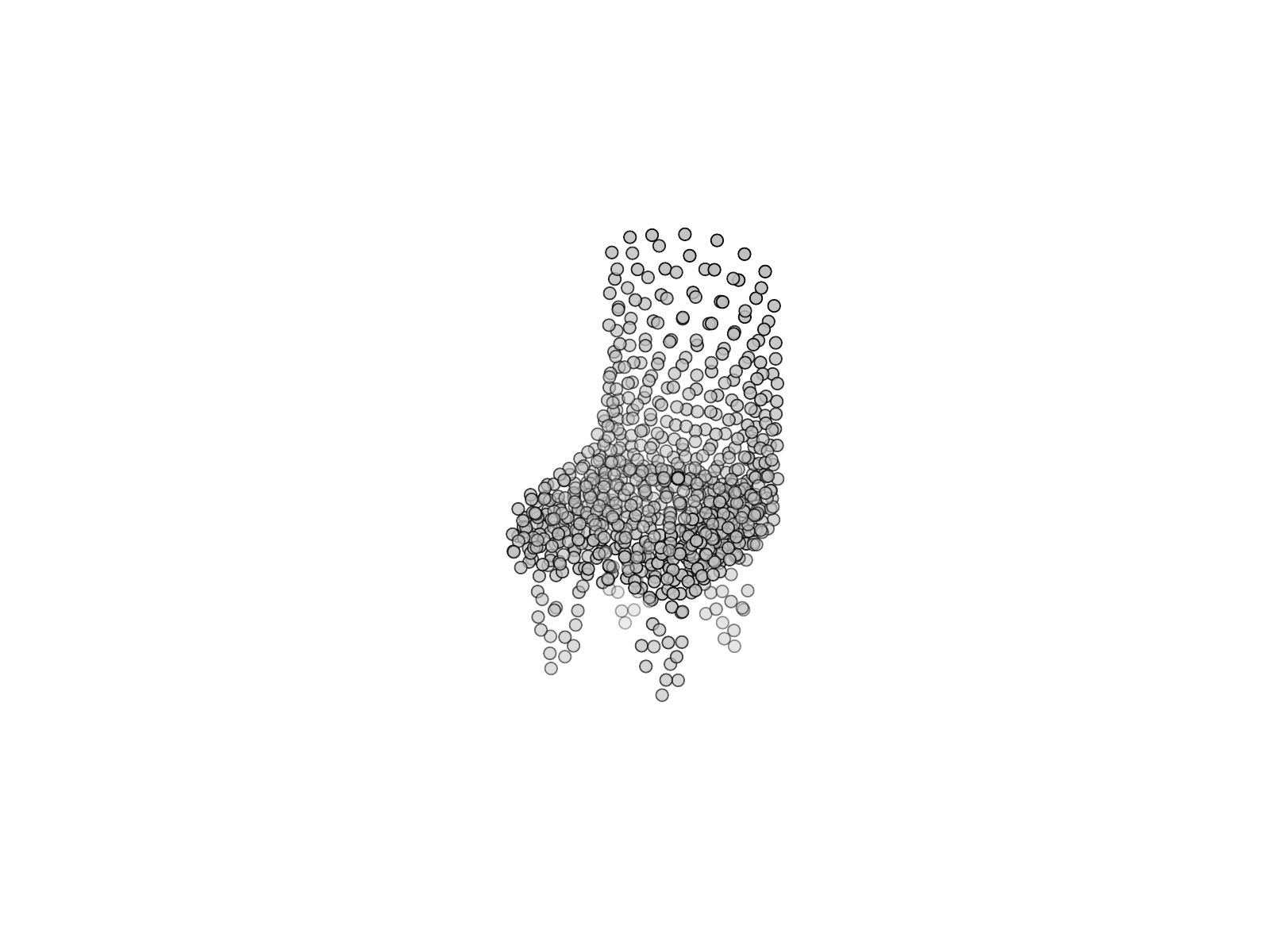}
\end{subfigure}\\
\begin{subfigure}[b]{\TextWidthPer\textwidth}
\includegraphics[trim={\TrLen cm \TrLen cm \TrLen cm \TrLen cm},clip,width=\textwidth]{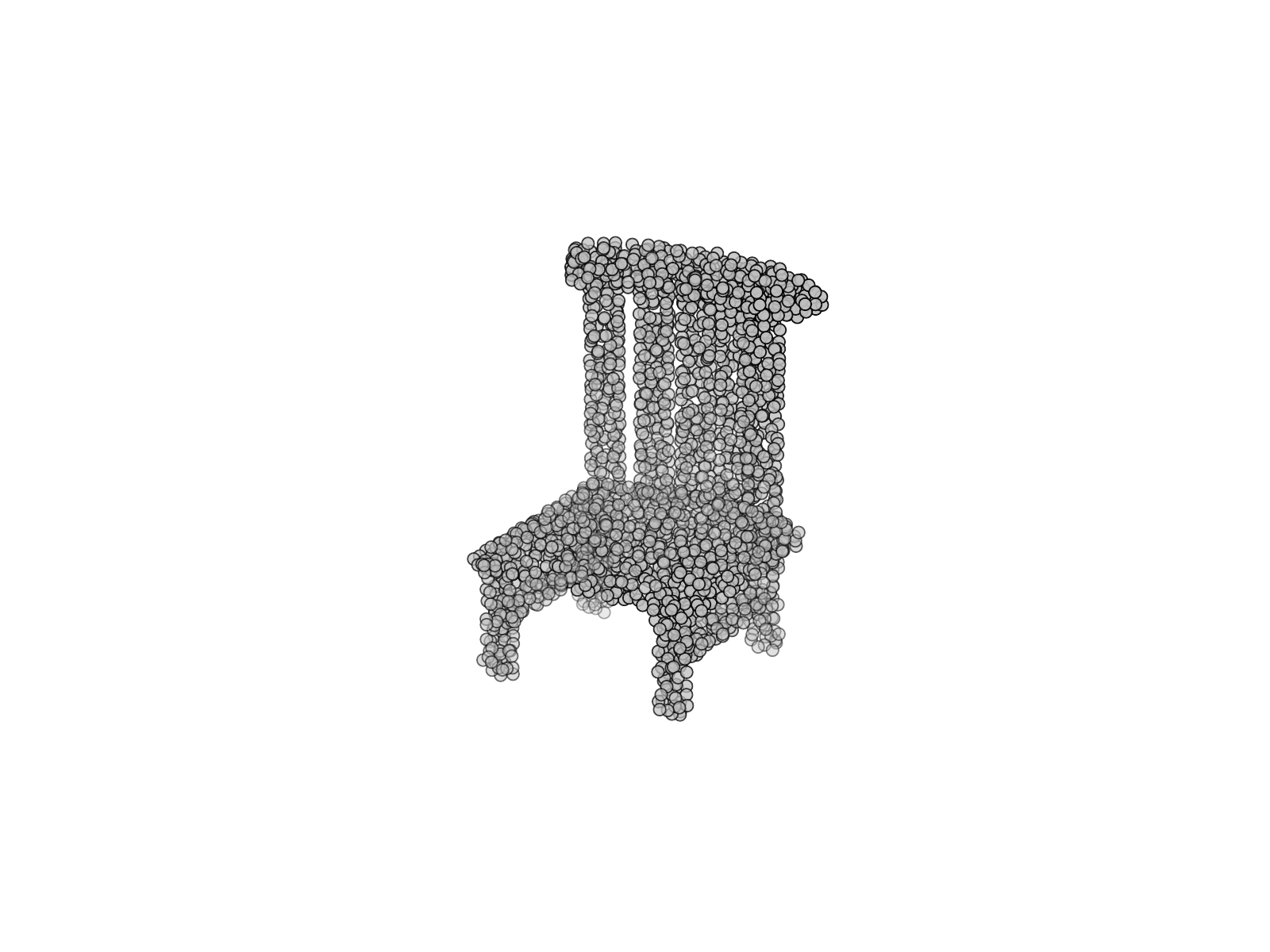}
\end{subfigure}
\begin{subfigure}[b]{\TextWidthPer\textwidth}
\includegraphics[trim={\TrLen cm \TrLen cm \TrLen cm \TrLen cm},clip,width=\textwidth]{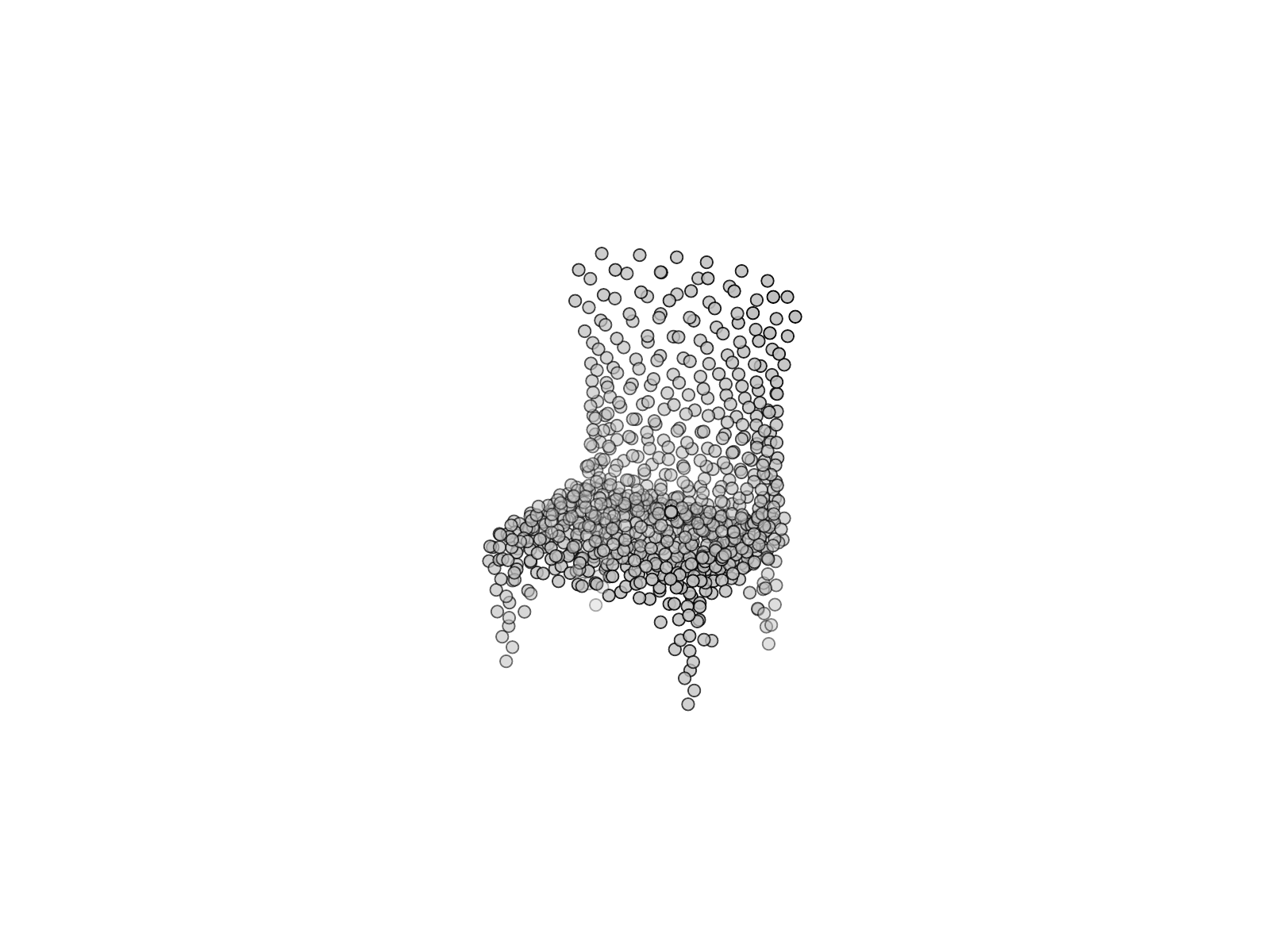}
\end{subfigure}
\begin{subfigure}[b]{\TextWidthPer\textwidth}
\includegraphics[trim={\TrLen cm \TrLen cm \TrLen cm \TrLen cm},clip,width=\textwidth]{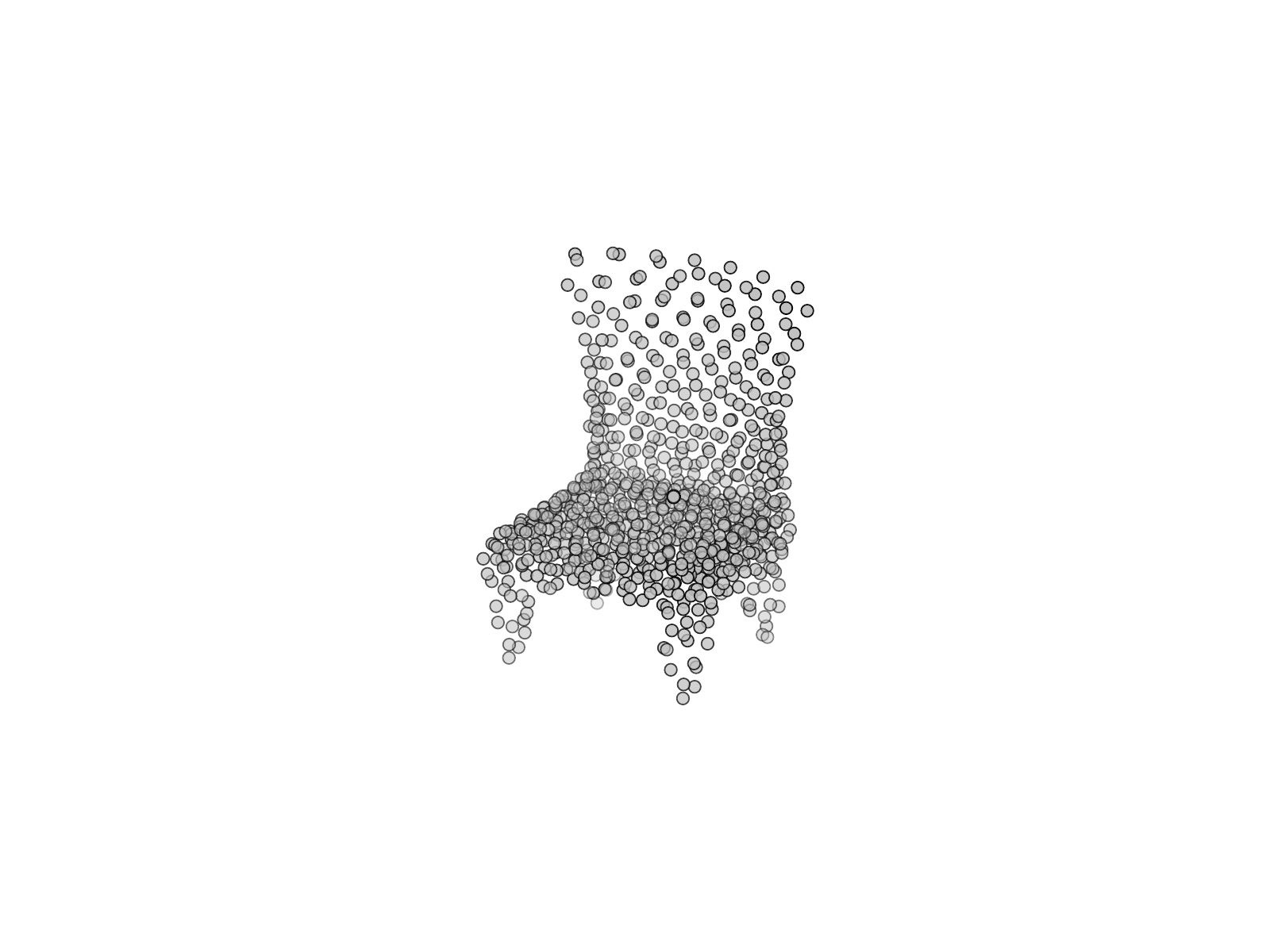}
\end{subfigure}\\
\begin{subfigure}[b]{\TextWidthPer\textwidth}
\includegraphics[trim={\TrLen cm \TrLen cm \TrLen cm \TrLen cm},clip,width=\textwidth]{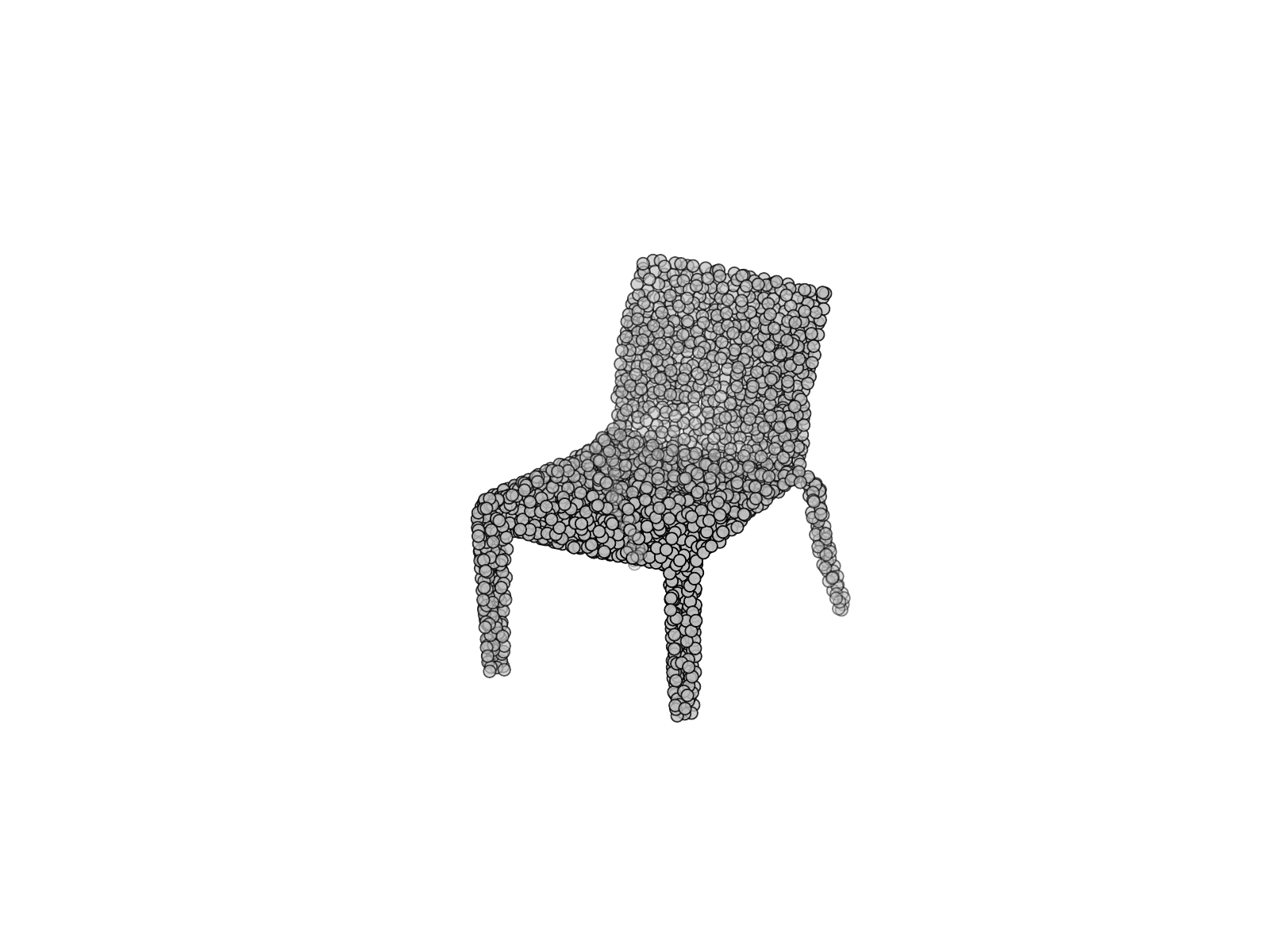}
\end{subfigure}
\begin{subfigure}[b]{\TextWidthPer\textwidth}
\includegraphics[trim={\TrLen cm \TrLen cm \TrLen cm \TrLen cm},clip,width=\textwidth]{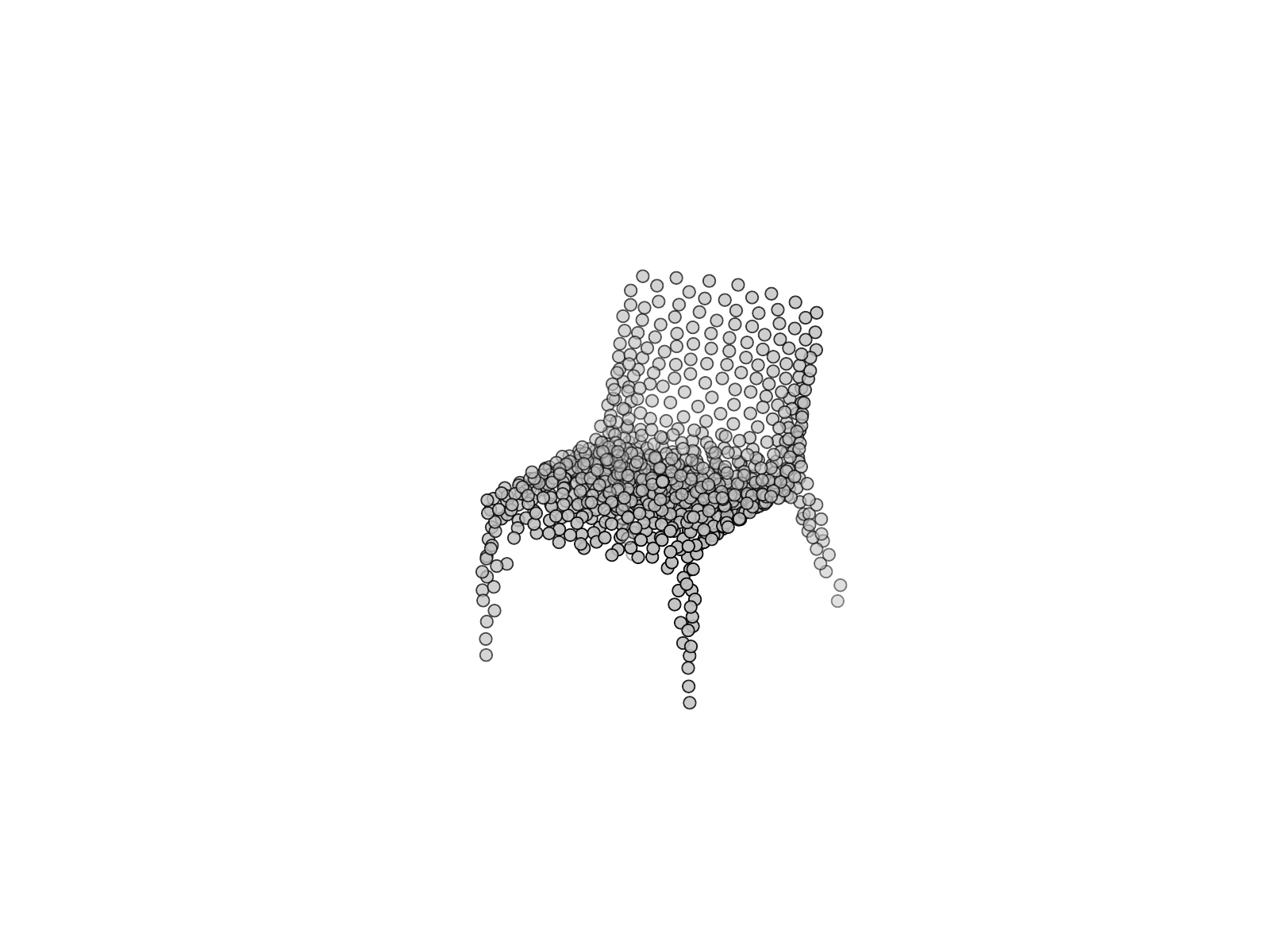}
\end{subfigure}
\begin{subfigure}[b]{\TextWidthPer\textwidth}
\includegraphics[trim={\TrLen cm \TrLen cm \TrLen cm \TrLen cm},clip,width=\textwidth]{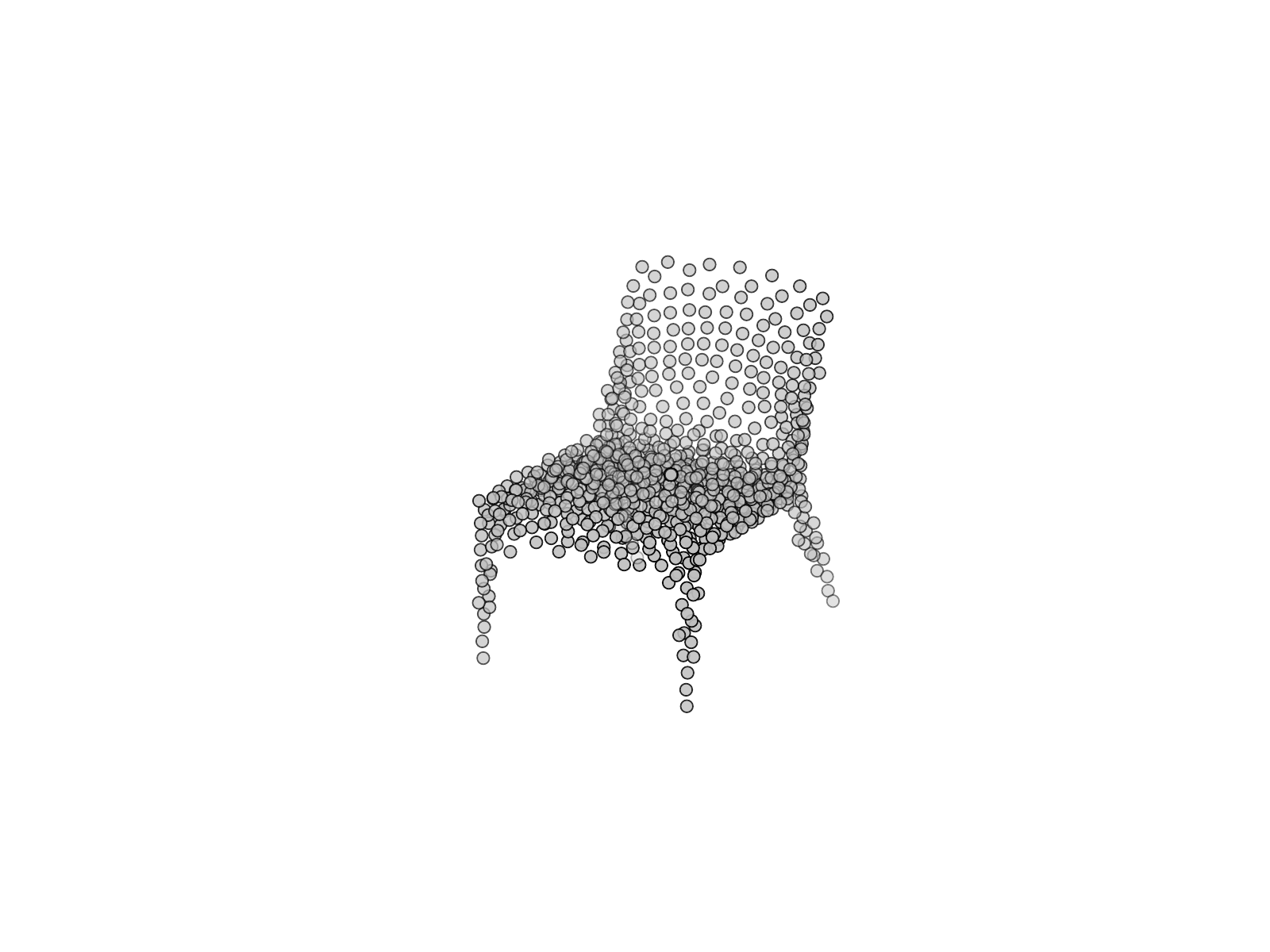}
\end{subfigure}\\
\begin{subfigure}[b]{\TextWidthPer\textwidth}
\includegraphics[trim={\TrLen cm \TrLen cm \TrLen cm \TrLen cm},clip,width=\textwidth]{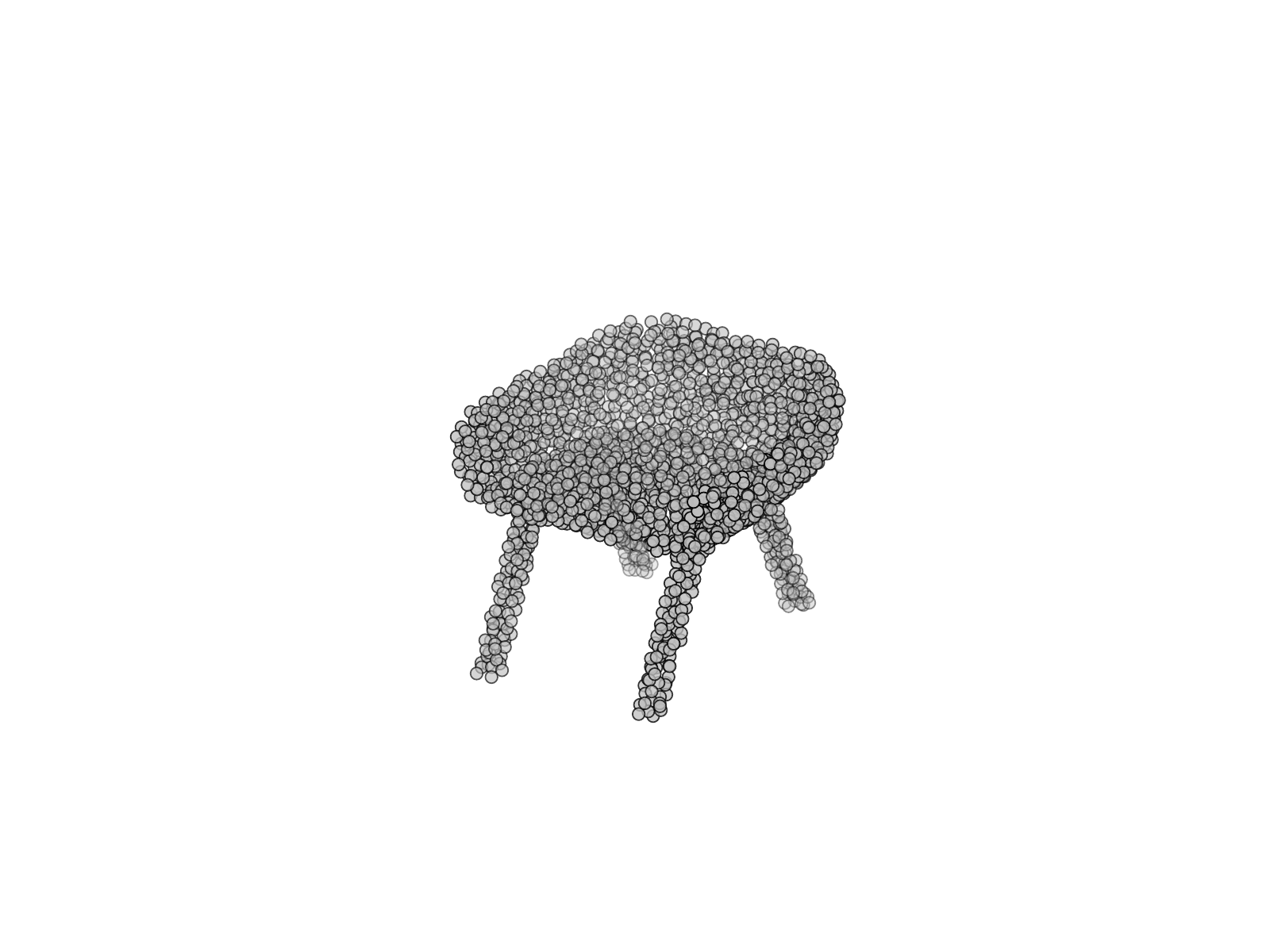}
\end{subfigure}
\begin{subfigure}[b]{\TextWidthPer\textwidth}
\includegraphics[trim={\TrLen cm \TrLen cm \TrLen cm \TrLen cm},clip,width=\textwidth]{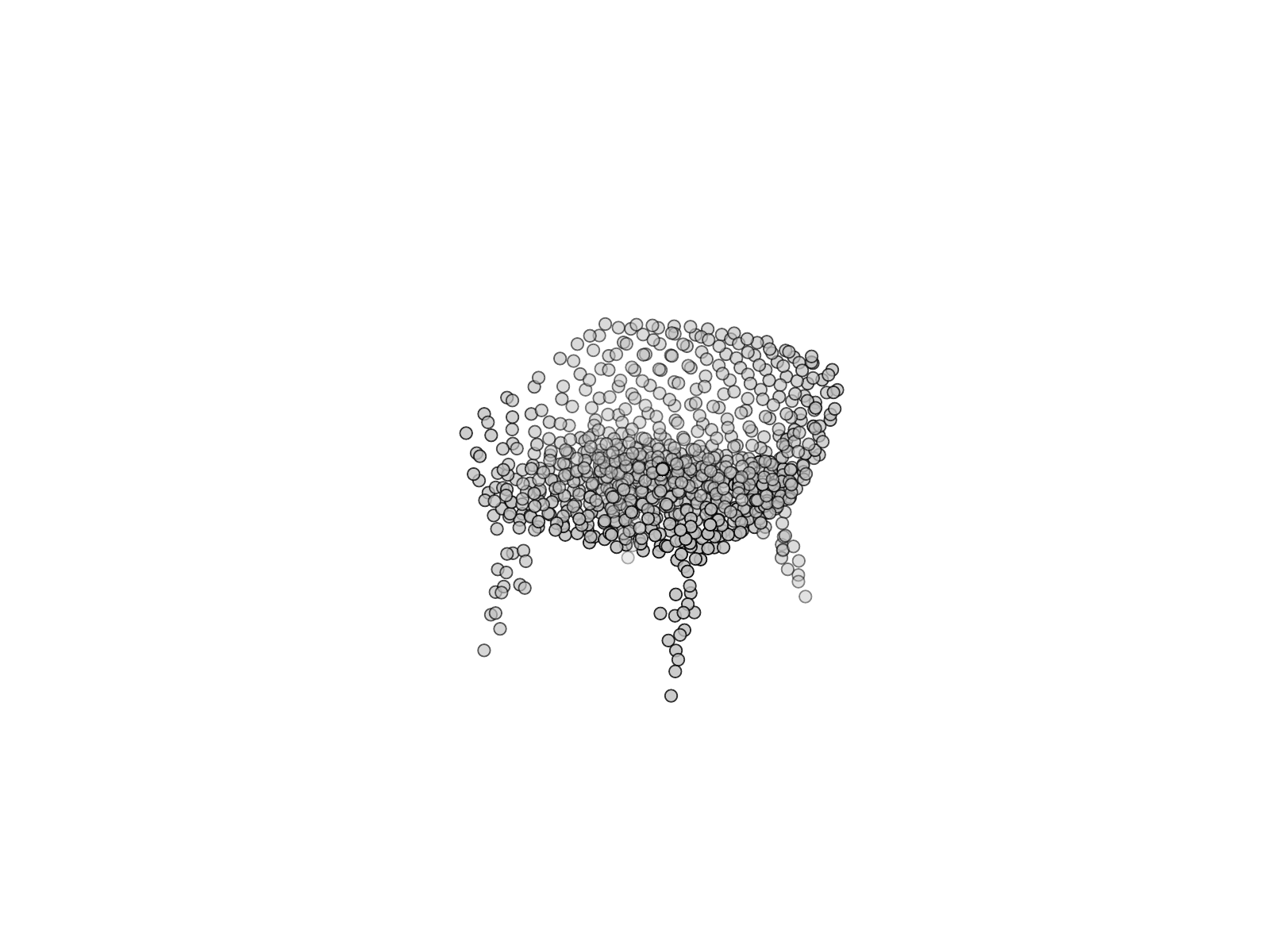}
\end{subfigure}
\begin{subfigure}[b]{\TextWidthPer\textwidth}
\includegraphics[trim={\TrLen cm \TrLen cm \TrLen cm \TrLen cm},clip,width=\textwidth]{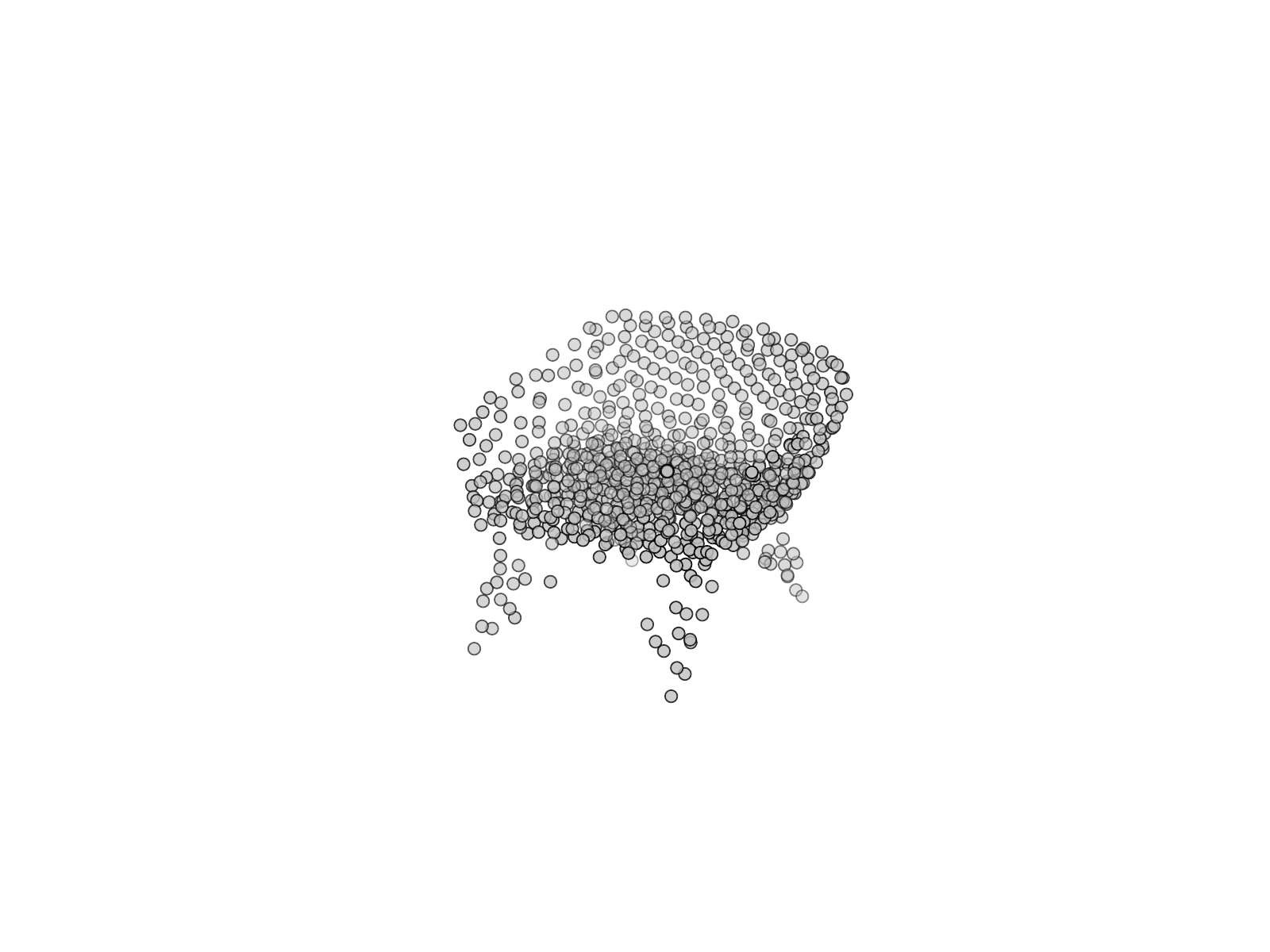}
\end{subfigure}\\
\caption{Autoencoder reconstructions for the \textsc{Chair} category from ShapeNet. Column 1: Ground truth, Column 2: Fully trained PointNet autoencoder, Column 3: Random encoder + trained decoder. }
\label{fig:autoencoder_vis}
\end{figure}
Results shown in Figure~\ref{fig:autoencoder_vis} suggest that random embeddings well capture the global structure of the shape.

\paragraph{t-SNE on Random Embeddings}
We choose $500$ random data instances from the ModelNet10 and MNIST test set, compute their random embeddings with PointNet-IN, and reduce their dimensionality to 2D using t-SNE~\cite{maaten2008:tsne} (perplexity $30$, iterations: $1000$, learning rate: $100.0$).
\begin{figure}
\centering
\begin{subfigure}[b]{0.3\textwidth}
\includegraphics[trim={1.25cm 1.25cm 1.25cm 1.25cm},clip,width=\textwidth]{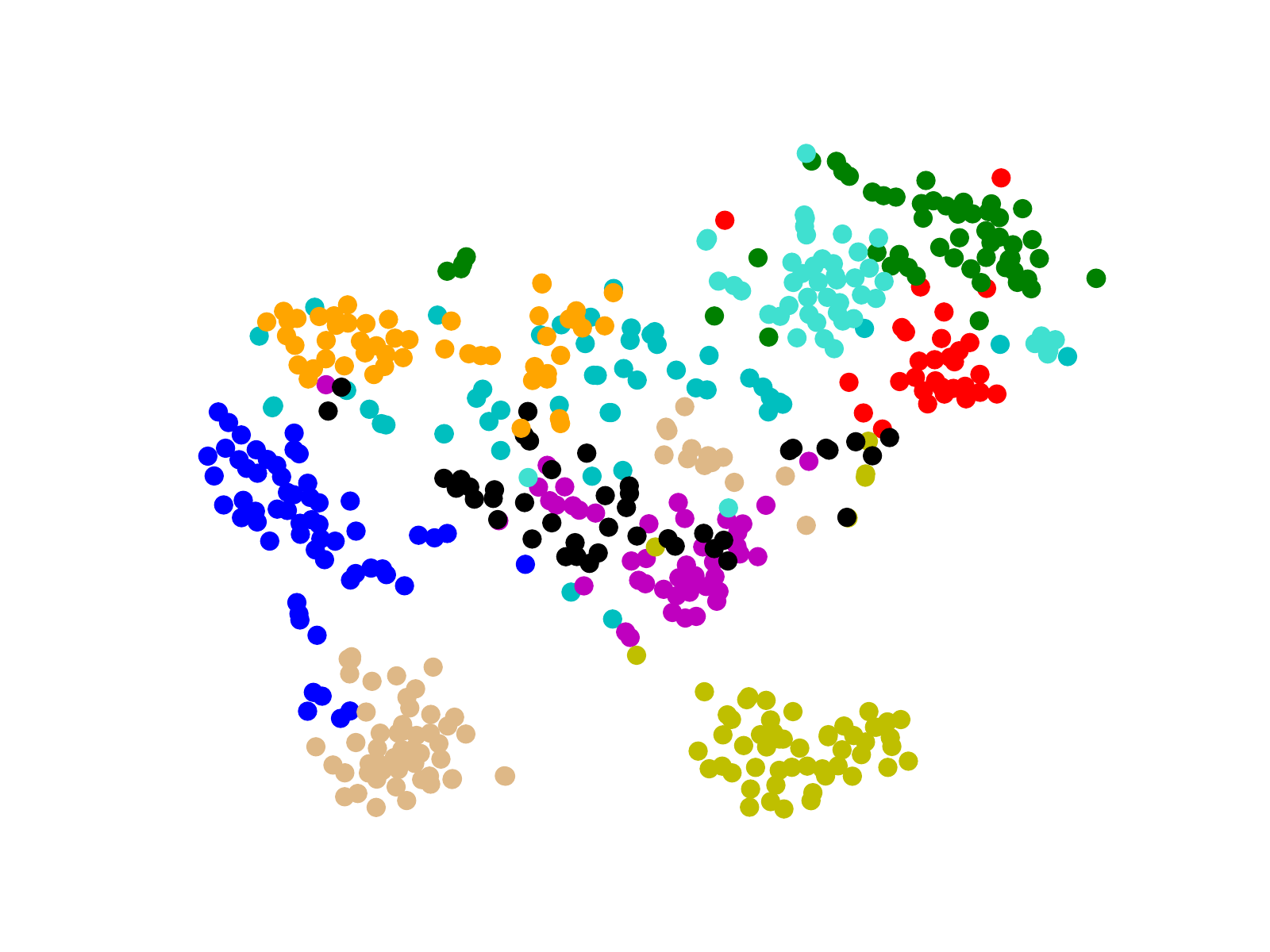}
\caption{ModelNet10}
\end{subfigure}
\begin{subfigure}[b]{0.3\textwidth}
\includegraphics[trim={1.5cm 1.5cm 1.5cm 1.5cm},clip,width=\textwidth]{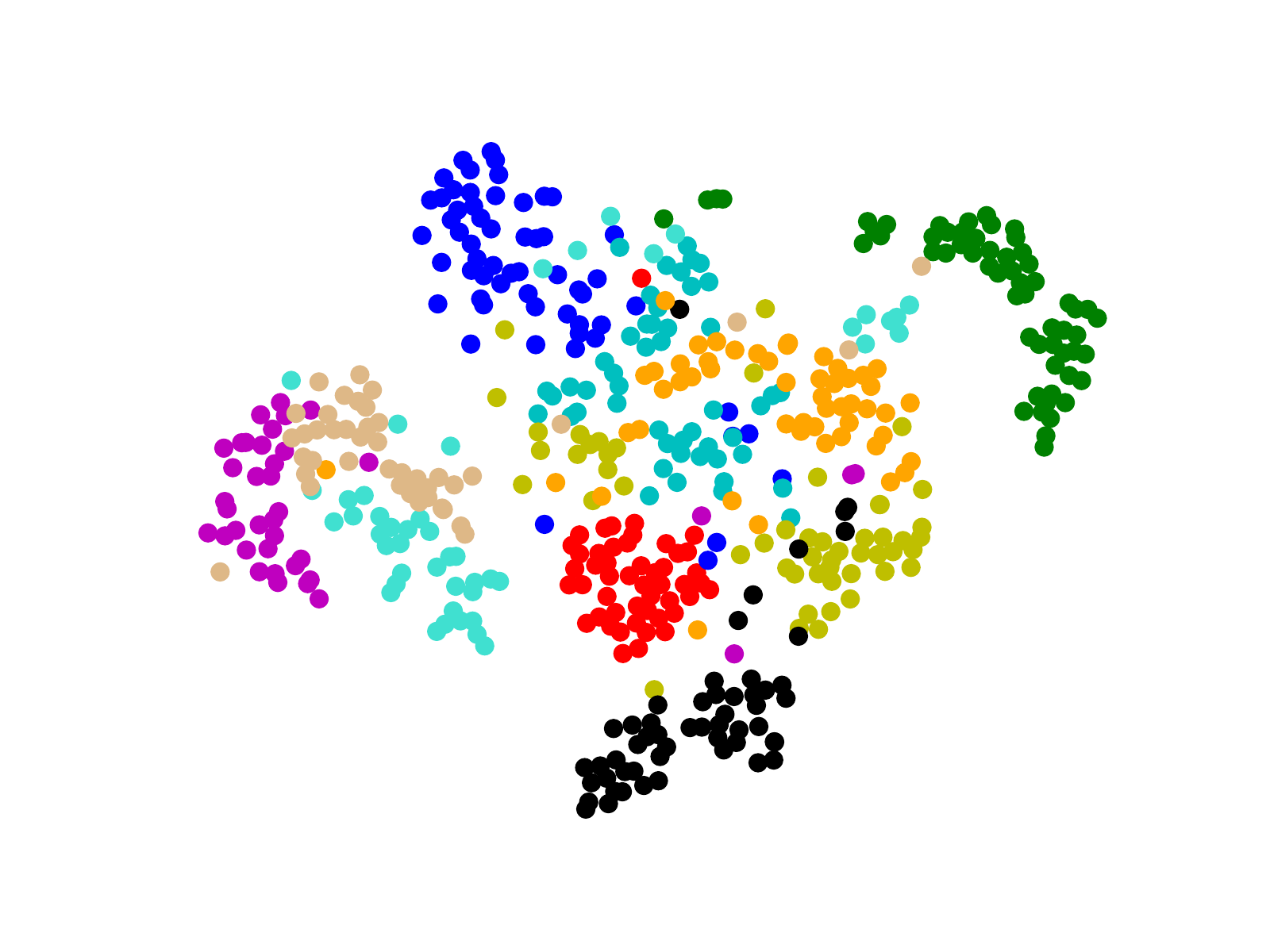}
\caption{MNIST}
\end{subfigure}
\caption{t-SNE visualization of random embeddings.}
\label{fig:tsne_vis}
\end{figure}
The results in Figure~\ref{fig:tsne_vis}(d, e) indicate that random embeddings well preserve the clustering of original input features.

\section{Conclusion}
\label{sec:conclusion}
In this work, we empirically investigated the efficacy of random pointcloud set functions for downstream tasks.
We discovered that nuances in the architecture (i.e., number of layers, normalization) play a major role in the separability of the random embeddings.
We also found that random neural set functions perform well on tasks where the datasets are well-aligned.
We demonstrated that random embeddings retain significant information about the original point cloud using unsupervised reconstruction.

Random set functions have potential to be used as a strong baseline, or when faster training time is desirable.
It would be interesting to uncover theoretical justifications behind the representational power of random set functions in future, e.g., by generalizing Cover's theorem, and uncover techniques to make trained encoders learn better representations than random encoders.

\bibliographystyle{plainnat}
\bibliography{references}

\begin{thebibliography}{18}
\providecommand{\natexlab}[1]{#1}
\providecommand{\url}[1]{\texttt{#1}}
\expandafter\ifx\csname urlstyle\endcsname\relax
  \providecommand{\doi}[1]{doi: #1}\else
  \providecommand{\doi}{doi: \begingroup \urlstyle{rm}\Url}\fi

\bibitem[Achlioptas et~al.(2018)Achlioptas, Diamanti, Mitliagkas, and
  Guibas]{achlioptas2018:latentgan}
Panos Achlioptas, Olga Diamanti, Ioannis Mitliagkas, and Leonidas Guibas.
\newblock Learning representations and generative models for 3{D} point clouds.
\newblock In Jennifer Dy and Andreas Krause, editors, \emph{Proceedings of the
  35th International Conference on Machine Learning}, volume~80 of
  \emph{Proceedings of Machine Learning Research}, pages 40--49, 2018.

\bibitem[Ba et~al.(2016)Ba, Kiros, and Hinton]{ba2016layer}
Jimmy~Lei Ba, Jamie~Ryan Kiros, and Geoffrey~E Hinton.
\newblock Layer normalization.
\newblock \emph{arXiv preprint arXiv:1607.06450}, 2016.

\bibitem[{Cover}(1965)]{cover65:theorem}
T.~M. {Cover}.
\newblock Geometrical and statistical properties of systems of linear
  inequalities with applications in pattern recognition.
\newblock \emph{IEEE Transactions on Electronic Computers}, EC-14\penalty0
  (3):\penalty0 326--334, 1965.
\newblock \doi{10.1109/PGEC.1965.264137}.

\bibitem[Frankle and Carbin(2018)]{frankle2018lottery}
Jonathan Frankle and Michael Carbin.
\newblock The lottery ticket hypothesis: Finding sparse, trainable neural
  networks.
\newblock \emph{arXiv preprint arXiv:1803.03635}, 2018.

\bibitem[Gaier and Ha(2019)]{gaier2019:weightagnostic}
Adam Gaier and David Ha.
\newblock Weight agnostic neural networks.
\newblock \emph{CoRR}, abs/1906.04358, 2019.
\newblock URL \url{http://arxiv.org/abs/1906.04358}.

\bibitem[Glorot and Bengio(2010)]{Glorot2010:weightinit}
Xavier Glorot and Yoshua Bengio.
\newblock Understanding the difficulty of training deep feedforward neural
  networks.
\newblock In \emph{Proceedings of the Thirteenth International Conference on
  Artificial Intelligence and Statistics}, pages 249--256, 2010.

\bibitem[He et~al.(2015)He, Zhang, Ren, and Sun]{he2015:weightinit}
Kaiming He, Xiangyu Zhang, Shaoqing Ren, and Jian Sun.
\newblock Delving deep into rectifiers: Surpassing human-level performance on
  imagenet classification.
\newblock In \emph{Proceedings of the 2015 IEEE International Conference on
  Computer Vision (ICCV)}, ICCV '15, pages 1026--1034. IEEE Computer Society,
  2015.
\newblock ISBN 978-1-4673-8391-2.
\newblock \doi{10.1109/ICCV.2015.123}.
\newblock URL \url{http://dx.doi.org/10.1109/ICCV.2015.123}.

\bibitem[He et~al.(2016)He, Wang, and Hopcroft]{he2016:randomvis}
Kun He, Yan Wang, and John Hopcroft.
\newblock A powerful generative model using random weights for the deep image
  representation.
\newblock In \emph{Proceedings of the 30th International Conference on Neural
  Information Processing Systems}, NIPS'16, pages 631--639, 2016.
\newblock ISBN 978-1-5108-3881-9.
\newblock URL \url{http://dl.acm.org/citation.cfm?id=3157096.3157167}.

\bibitem[Ioffe and Szegedy(2015)]{ioffe2015batch}
Sergey Ioffe and Christian Szegedy.
\newblock Batch normalization: Accelerating deep network training by reducing
  internal covariate shift.
\newblock \emph{arXiv preprint arXiv:1502.03167}, 2015.

\bibitem[Maaten and Hinton(2008)]{maaten2008:tsne}
Laurens van~der Maaten and Geoffrey Hinton.
\newblock Visualizing data using t-sne.
\newblock \emph{Journal of machine learning research}, 9:\penalty0 2579--2605,
  2008.

\bibitem[Qi et~al.(2016)Qi, Su, Mo, and Guibas]{qi2016:pointnet}
Charles~Ruizhongtai Qi, Hao Su, Kaichun Mo, and Leonidas~J. Guibas.
\newblock Pointnet: Deep learning on point sets for 3d classification and
  segmentation.
\newblock \emph{CoRR}, abs/1612.00593, 2016.
\newblock URL \url{http://arxiv.org/abs/1612.00593}.

\bibitem[Ulyanov et~al.(2016)Ulyanov, Vedaldi, and
  Lempitsky]{Ulyanov2016:instancenorm}
Dmitry Ulyanov, Andrea Vedaldi, and Victor~S. Lempitsky.
\newblock Instance normalization: The missing ingredient for fast stylization.
\newblock \emph{CoRR}, abs/1607.08022, 2016.
\newblock URL \url{http://arxiv.org/abs/1607.08022}.

\bibitem[Ulyanov et~al.(2017)Ulyanov, Vedaldi, and
  Lempitsky]{ulyanov2017:deepimageprior}
Dmitry Ulyanov, Andrea Vedaldi, and Victor~S. Lempitsky.
\newblock Deep image prior.
\newblock \emph{CoRR}, abs/1711.10925, 2017.
\newblock URL \url{http://arxiv.org/abs/1711.10925}.

\bibitem[Wagstaff et~al.(2019)Wagstaff, Fuchs, Engelcke, Posner, and
  Osborne]{wagstaff2019limitations}
Edward Wagstaff, Fabian~B Fuchs, Martin Engelcke, Ingmar Posner, and Michael
  Osborne.
\newblock On the limitations of representing functions on sets.
\newblock \emph{arXiv preprint arXiv:1901.09006}, 2019.

\bibitem[Wieting and Kiela(2019)]{wieting2019:randomword}
John Wieting and Douwe Kiela.
\newblock No training required: Exploring random encoders for sentence
  classification.
\newblock \emph{CoRR}, abs/1901.10444, 2019.
\newblock URL \url{http://arxiv.org/abs/1901.10444}.

\bibitem[Wu et~al.(2016)Wu, Zhang, Xue, Freeman, and
  Tenenbaum]{jiajun2016:3dgan}
Jiajun Wu, Chengkai Zhang, Tianfan Xue, William~T Freeman, and Joshua~B
  Tenenbaum.
\newblock Learning a probabilistic latent space of object shapes via 3d
  generative-adversarial modeling.
\newblock In \emph{Advances in Neural Information Processing Systems}, pages
  82--90, 2016.

\bibitem[Yang et~al.(2018)Yang, Feng, Shen, and Tian]{yang2018:foldingnet}
Yaoqing Yang, Chen Feng, Yiru Shen, and Dong Tian.
\newblock Foldingnet: Point cloud auto-encoder via deep grid deformation.
\newblock In \emph{Proceedings of the IEEE Conference on Computer Vision and
  Pattern Recognition}, pages 206--215, 2018.

\bibitem[Zaheer et~al.(2017)Zaheer, Kottur, Ravanbakhsh, P{\'{o}}czos,
  Salakhutdinov, and Smola]{zaheer2017:deepsets}
Manzil Zaheer, Satwik Kottur, Siamak Ravanbakhsh, Barnab{\'{a}}s P{\'{o}}czos,
  Ruslan Salakhutdinov, and Alexander~J. Smola.
\newblock Deep sets.
\newblock \emph{CoRR}, abs/1703.06114, 2017.
\newblock URL \url{http://arxiv.org/abs/1703.06114}.

\end{thebibliography}

\end{document}